\documentclass{article}

\PassOptionsToPackage{numbers}{natbib}
\usepackage[main, final]{neurips_2026}

\usepackage[utf8]{inputenc} \usepackage[T1]{fontenc}    \usepackage{hyperref}       \usepackage{url}            \usepackage{booktabs}       \usepackage{amsfonts}       \usepackage{nicefrac}       \usepackage{microtype}      \usepackage{xcolor}         \usepackage{graphicx}
\usepackage{subcaption}
\usepackage{amsmath, amssymb, amsthm, graphicx, mathtools, enumitem, hyperref, booktabs, tikz, pgfplots, float, subcaption, adjustbox, bbm, caption, multirow}

\usepackage{url}            \usepackage{amsfonts}       \usepackage{nicefrac}       \usepackage{microtype}

\usepackage{array}
\usepackage{varwidth}
\usepackage{wrapfig}
\usepackage{algorithm}
\usepackage{algorithmic}
\usepackage{footnote}
\makesavenoteenv{tabular}
\makesavenoteenv{table}
\usepackage[flushleft]{threeparttable}
\usepackage{color}
\usepackage{soul}
\usepackage{booktabs}
\usepackage{enumitem}
\usepackage{tablefootnote}
\usepackage{bm}
\usepackage[capitalize,noabbrev]{cleveref}
\usepackage[table,xcdraw,dvipsnames]{xcolor}
\newcolumntype{C}[1]{>{\centering\arraybackslash}m{#1}}
\usepackage[capitalize,noabbrev]{cleveref}

\theoremstyle{plain}
\newtheorem{theorem}{Theorem}[section]
\newtheorem{proposition}[theorem]{Proposition}

\theoremstyle{definition}

\theoremstyle{remark}
\newtheorem{remark}[theorem]{Remark}

\newcommand{\D}{\mathcal D}

\newcommand{\Y}{\mathcal Y}

\title{Generative Conformal Prediction with\\ Optimized Coverage Allocation}

\author{Minxing Zheng \\
  Carnegie Mellon University \\
  \And
  Shixiang Zhu \\
  Carnegie Mellon University \\
}

\begin{document}

\maketitle

\begin{abstract}
  Conformal prediction provides model-agnostic uncertainty quantification with guaranteed coverage, but conventional methods often yield overly conservative uncertainty sets, particularly in multimodal or heterogeneous settings. This inefficiency arises from two sources: ($i$) limited expressiveness of the predictive model and ($ii$) simplistic nonconformity scores design. Most existing approaches advance only one of these axes, leaving the other underexplored.
We propose generative conformal prediction with Optimized Ranking and Coverage Allocation (\texttt{ORCA}), a three-stage framework that advances both aspects jointly. \texttt{ORCA} leverages generative models to capture the full conditional distribution and introduces a rank-dependent optimization procedure that adaptively allocates coverage for efficiency while maintaining validity. We cast this coverage allocation as an optimization problem, derive an exact mixed-integer linear programming formulation, and show that the solution converges asymptotically to the oracle density-level set.
Across synthetic, semi-synthetic, and real datasets, \texttt{ORCA} produces substantially more efficient uncertainty sets than state-of-the-art baselines, demonstrating robust gains in scenarios where conventional conformal prediction methods fail.

\end{abstract}

\vspace{-.1in}
\section{Introduction}
\vspace{-.1in}

Conformal prediction (CP) \citep{Vovk:2005} has gained prominence as a powerful framework for uncertainty quantification, offering uncertainty sets with statistically guaranteed coverage under minimal assumptions.
As a model-agnostic approach, CP can enhance any black-box model to provide uncertainty sets that contain the ground truth with a user-specified probability. This versatility has driven its adoption across diverse domains, including healthcare \citep{Seoni:2023}, finance \citep{Fujimoto:2022,Blasco:2024}, and autonomous systems \citep{Michelmore:2020,Su:2023,Grewal:2024}, where reliable decision-making under uncertainty is critical.

Despite its strengths in coverage validity and model-agnostic applicability, CP often yields overly conservative uncertainty sets, limiting practicality and interpretability when the conditional distribution $p(Y \mid X)$ exhibits complex structure such as multimodality or heterogeneity. This inefficiency arises from two sources: ($i$) limited model capacity to capture the geometry of the conditional distribution, and ($ii$) simplistic nonconformity scores that fail to exploit distributional richness. For instance, linear regression provides only conditional mean estimates, whereas quantile regression could recover multiple conditional quantiles and thus reveals more of the distribution. Likewise, for a regression model, one may choose absolute residual as the score, or a more sophisticated variant such as the locally weighted nonconformity measure that normalizes residuals to improve efficiency \citep{Papadopoulos:2002}. Both directions of using more expressive models and designing more refined scores enhance uncertainty quantification.
Recent work has pushed along these axes; however, most approaches treat model expressiveness and score design in isolation, advancing one while neglecting the other. In this work, we unify these perspectives by proposing a framework that couples generative modeling with rank-dependent, vectorized score constructions optimized for efficient coverage allocation.

\begin{figure}[!t]
\centering
\includegraphics[width=\linewidth]{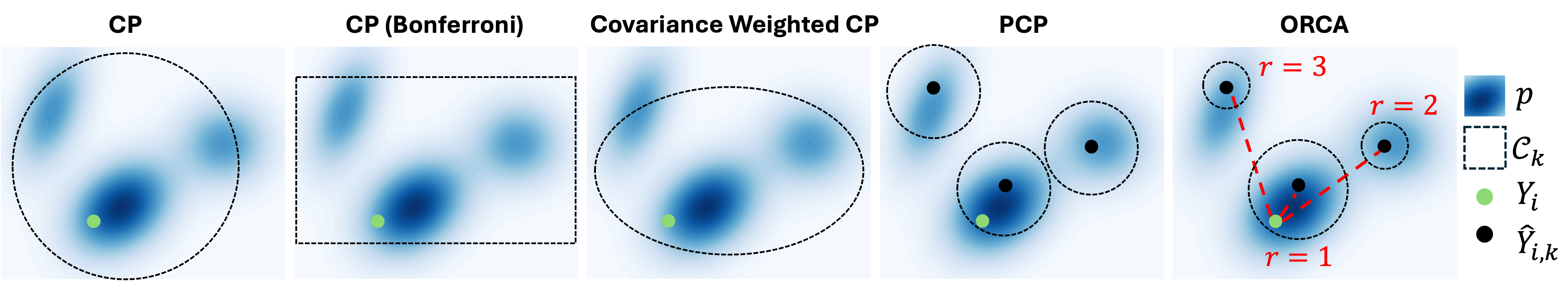}
\vspace{-.2in}
\caption{Comparison of conformal prediction methods and their prediction regions. From left to right: standard CP \citep{Vovk:2005}, Bonferroni-corrected CP \citep{Neeven:2018}, Covariance Weighted CP \citep{Messoudi:2022,Xu:2024}, PCP \citep{wang:2023}, and \texttt{ORCA} (our method). The background gradient shows the multimodal data distribution. \texttt{ORCA} adjusts circle radii via ranking—larger in dense areas, smaller in sparse ones—yielding more efficient and flexible prediction sets. Green dots denote true responses ($Y_i$), black dots are generated samples ($\hat{Y}_{i,k}$), and dashed lines mark prediction boundaries ($\mathcal{C}_k$).
}
\label{fig:non-conformity-score}
\vspace{-.1in}
\end{figure}

In this paper, we propose generative conformal prediction with Optimized Ranking and Coverage Allocation (\texttt{ORCA}), a three-stage framework that combines generative modeling with optimization-based, efficiency-oriented non-conformity score design. \texttt{ORCA} leverages generative samples to capture the full conditional distribution and introduces a rank-dependent optimization procedure that adaptively allocates coverage while maintaining exact finite-sample validity. This yields prediction sets that adapt to the geometry of $Y \mid X$, improving efficiency. Figure~\ref{fig:non-conformity-score} highlights our advances over classical conformal prediction baselines, illustrating how \texttt{ORCA} achieves a more adaptive uncertainty set. In summary, the main contributions are:

\begin{enumerate}[left=0pt,topsep=0pt,parsep=0pt,itemsep=2pt]
    \item \emph{Generative conformal prediction.} \texttt{ORCA} employs generative models as predictors to capture distributional complexity beyond mean- or quantile-based approaches. By leveraging generative sampling with local density proxy ranking mechanism, \texttt{ORCA} exploits the fitted conditional distribution, encoding finer distributional structure into the conformal procedure.

    \item \emph{Optimized coverage allocation.} Rather than collapsing generated sample residuals into a single statistic, \texttt{ORCA} operates on rank-dependent residual vectors. We cast conformal coverage as an explicit coverage allocation optimization problem and reformulate it as a mixed-integer linear program (MILP), which ensures global optimality with practical scalability.

    \item \emph{Theory with oracle guarantees.} We prove that \texttt{ORCA} maintains exact finite-sample validity and, asymptotically, converges to the oracle optimal highest-density region, thereby establishing a rigorous link between optimization-based conformal prediction and density-level sets.

    \item \emph{Empirical evaluation.} Experiments on synthetic and real datasets show that \texttt{ORCA} achieves more efficient uncertainty sets than baselines while maintaining coverage.
\end{enumerate}

\vspace{-.1in}
\paragraph{Related Work}

Extended works have been done focusing on CP for deterministic models. Different summary statistics of the conditional distribution have been studied, for example, mean \citep{Shafer:2008,Lei:2018} and quantiles \citep{Romano:2019,Kivaranovic:2020,Alaa:2023}.
\cite{Sesia:2021} estimate conditional distributions with histograms to obtain efficient intervals with approximate conditional coverage. Though the above methods all provide continuous prediction intervals, they cannot adapt to multimodality well. In \cite{Lei:2013} and \cite{Lei:2014}, the authors first consider the aspect of density level sets and density-based clustering for providing efficient prediction set. The authors use kernel density estimator to construct the non-conformity scores and with a proper choice of bandwidth, the prediction set could be discontinuous.
There are also following works \cite{Izbicki:2020,Chernozhukov:2021,Izbicki:2022} that estimate the density or cumulative distribution function to construct prediction sets. As a reminder, in our setting, we only need access to the random samples generated from the fitted model instead of explicit density estimation. So our setting is more generic and applicable to real applications. Other than density estimation, \cite{Tumu:2024} fits shape functions (convex hull, hyperrectangle, and ellipsoid) on the calibration data to provide a discontinuous prediction set as a union of fitted shapes. In \cite{Guha:2024}, the authors convert regression to a classification problem and can provide discontinuous prediction sets with interpolation of classification scores.

Efficiency, measured by the size of the prediction set, is a fundamental assessment of the quality of a CP method. Some works explore the effect of different non-conformity score measures such as by scaling and reweighting the non-conformity score of each calibration data \citep{Papadopoulos:2002,Papadopoulos:2011,Kivaranovic:2020,Bellotti:2020,Amoukou:2023}.
On the other hand, some works \citep{stutz2021learning,Einbinder:2022} directly reduce inefficiency by incorporating a differentiable inefficiency measurement into the training loss. \cite{bai2022efficient} reformulate CP as a constrained optimization problem. By introducing an extra parameter and enriching the search space of the optimization problem, it demonstrates better performance in efficiency. Our main idea is partially motivated by this reformulation.
We also gain insights from the recent development of CP for multi-target regression tasks and the definition of multivariate quantile function as discussed in \citep{Messoudi:2021,sun:2022}.
Recent advancements, such as probabilistic conformal prediction (PCP) using conditional random samples \citep{wang:2023}, have paved the way for CP in generative models. PCP leverages generated random samples to construct the uncertainty sets in the form of a union of balls around each generated sample, offering a discontinuous uncertainty set. This is a fundamental step towards better coverage efficiency, as ideally, we want the prediction set to avoid low-density regions.
However, PCP faces two major limitations: ($i$) PCP does not distinguish different generated samples and assigns the same radius to each uncertainty ball, ($ii$) when the number of generated samples is relatively small, PCP can yield conservative prediction sets.

\vspace{-.1in}
\section{Preliminaries}
\vspace{-.1in}

Conformal Prediction (CP) is a general framework for uncertainty quantification that provides statistically valid prediction regions for any predictive model under the mild assumption of data exchangeability \cite{Shafer:2008,angelopoulos2023conformal,Tibshirani:2019}. For any test input $X_{\text{test}}$, CP aims to construct a prediction set $\hat{\mathcal{C}}(X_{\text{test}})$ that satisfies the coverage guarantee
\begin{equation}
    \label{eq:cp_guarantee}
    \text{Pr}\bigl(Y_{\text{test}} \in \hat{\mathcal{C}}(X_{\text{test}})\bigr) \geq 1 - \alpha,
\end{equation}
for a user-specified miscoverage rate $\alpha \in (0,1)$.

Given calibration dataset $\D = \{(X_i, Y_i)\}_{i=1}^n$ with $X_i \in \mathcal{X}$ are features and $Y_i \in \mathcal{Y}$ are responses, CP defines a \emph{non-conformity score} $E_i$ that measures how each sample conforms to fitted model $\hat{f}$. In regression, a common choice is the prediction error $ E_i = \| Y_i - \hat{f}(X_i) \|.$
The $\lceil (n+1)(1-\alpha) \rceil$-th order statistics of the calibration scores $\{E_i\}$ yields a threshold $\tau$, and the prediction set is
\[
    \hat{\mathcal{C}}(X_{\text{test}}) = \{y : \|y - \hat{f}(X_{\text{test}})\| \leq \tau \}.
\]
The validity of \eqref{eq:cp_guarantee} requires only that $(X_{\text{test}}, Y_{\text{test}})$ be exchangeable with $\D$.

\noindent\emph{Prediction Set Efficiency.}
Beyond validity, \emph{efficiency}, the size of $\hat{\mathcal{C}}(X_{\text{test}})$, is crucial: trivial sets covering all of $\mathcal{Y}$ achieve coverage but are uninformative. We therefore seek prediction sets that are valid and as small as possible.

\vspace{-.1in}
\section{Proposed Method}
\vspace{-.1in}

In this section, we introduce generative conformal prediction with Optimized Ranking and Coverage Allocation (\texttt{ORCA}), a framework that leverages generative models to construct efficient uncertainty sets with finite-sample coverage guarantees. Unlike classical conformal prediction methods that rely on point estimates or low-dimensional summaries, generative models provide samples from the full conditional distribution $\hat p(Y\mid X)$, enabling richer characterization of predictive uncertainty. Existing generative conformal approaches typically reduce these samples to a single nonconformity statistic, which discards important distributional structure and often leads to inefficient prediction sets.

Our key idea is to preserve and exploit the geometric structure of the generated samples through a vectorized representation of sample distances. In particular, we rank generated samples according to their local density and construct uncertainty sets as a union of rank-dependent regions, allowing different parts of the predictive distribution to receive different coverage allocations. Formally, the resulting prediction set takes the form
\begin{equation}
\label{eq:coverage_set_init}
    \hat{\mathcal{C}}_{\textrm{ORCA}}(X_{\text{test}})
    = \bigcup_{r=1}^K {\mathcal{B}}_r(X_{\text{test}};R_r(\alpha)),
\end{equation}
where $R(\alpha)\in\mathbb{R}^K$ is a vector of rank-dependent radii, and  ${\mathcal{B}}_r(X_{\text{test}};R_r(\alpha))$ denotes a region centered at the rank $r$ generated sample with radius $R_r(\alpha)$. When $Y\in\mathbb{R}$, these regions reduce to intervals, while for $Y\in\mathbb{R}^d$ they correspond to Euclidean balls, but they can also be general metric in ourput space.

The proposed framework combines three ingredients: ($i$) distribution exploration through generative sampling and density-aware ranking, ($ii$) optimization-based allocation of coverage across ranked regions to maximize efficiency, and ($iii$) conformal calibration to guarantee finite-sample marginal coverage.
These components allow \texttt{ORCA} to approximate oracle density level sets while retaining the rigorous validity guarantees of conformal prediction.

\begin{algorithm}[t]
    \centering
    \caption{\textbf{ORCA}}
    \label{alg:gcpvcr}
    \small
    \begin{algorithmic}[1]
        \STATE \textbf{Input}: generative model $\hat{p}(Y|X)$; splits $\D_{\mathrm{explore}},\D_{\text{calib}}$; test feature $X_{\text{test}}$; miscoverage $\alpha$; sample size $K$
        \STATE \textbf{Stage~1: Exploration with Sampling and Ranking}
        \FOR{$(X_i,Y_i)\in\D_{\mathrm{explore}}$}
            \STATE Sample $\{\hat Y_{i,k}\}_{k=1}^K\sim\hat p(\cdot|X_i)$
            \STATE Rank samples by $m$-NN density (Eq.~\eqref{eq:ranking})
            \STATE Construct distance vector $E_i$ via Eq.~\eqref{eq:rankpcp_score}
        \ENDFOR
        \STATE \textbf{Stage~2: Quantile Optimization}
        \STATE Solve optimization problem \eqref{eq:optim_problem} (via MILP \eqref{eq:milp}) to obtain threshold $Q_r(\hat\beta_r)$ for each $r$.
        \STATE \textbf{Stage~3: Calibration}
        \FOR{$(X_i,Y_i)\in\D_{\text{calib}}$}
            \STATE Compute nonconformity scores $S_i$ (Eq. \eqref{eq:nonconformity_score})
        \ENDFOR
        \STATE Calibrate coverage threshold vector $R(\alpha)$ (Eq.\eqref{eq:coverage_threshold})
        \STATE \textbf{Prediction:} Draw $\{\hat Y_{\text{test},k}\}_{k=1}^K \sim \hat{p}(Y|X_{\text{test}})$ and rank, form $\hat{\mathcal{C}}_{\textrm{ORCA}}(X_{\text{test}})$ via Eq.~\eqref{eq:coverage_set}.
    \end{algorithmic}
\end{algorithm}

\vspace{-0.1in}
\subsection{Stage~1: Efficiency-Oriented Distributional Quantile Exploration}
\vspace{-0.1in}

The goal of Stage~1 is to explore the geometry of the predictive distribution. It produces rank-dependent distance vectors by sampling from the fitted model and ranking the generated samples by local density proxies. Intuitively, these vectors fully preserve distributional structure, including high- and low-density regions as well as multimodality.

We begin with a holdout dataset
$\mathcal{D}=\{(X_i,Y_i)\}_{i=1}^{|\mathcal D|}$, distinct from the training data, and partition it into two disjoint subsets: $\mathcal D=\mathcal D_{\mathrm{explore}}\cup\mathcal D_{\mathrm{calib}}$, with $|\mathcal D_{\mathrm{explore}}|=n_1$ and $|\mathcal D_{\mathrm{calib}}|=n_2$. We refer to these subsets as the \emph{exploration split} and the \emph{calibration split}, respectively.
For each data point $(X_i,Y_i) \in \mathcal{D}_{\text{explore}}$, we draw $K$ conditional samples $\{\hat Y_{i,k}\}_{k=1}^K \sim \hat p(\cdot \mid X_i)$. We compute average $m$-nearest-neighbor distance for each sample $\hat Y_{i,k}$, denoted $\bar D_{i,k}$, and rank the samples by distances:
\begin{equation}\label{eq:ranking}
\bar D_{i,j(1)} \le \cdots \le \bar D_{i,j(K)},
\end{equation}
where $j(r)$ indexes the sample with the $r$-th smallest $m$-nearest-neighbor distance. We form \emph{ranked distance vector}
\begin{equation}\label{eq:rankpcp_score}
E_i = (E_{i,1},\dots,E_{i,K}), \quad E_{i,r} = \|Y_i - \hat Y_{i,j(r)}\|,
\end{equation}
where each entry records the distance between the observed label $Y_i$ and the $r$-th ranked sample.

The rationale of ranking and constructing the ranked distance vector is two-fold:
$(i)$ By retaining a full vector rather than a single scalar, we capture a richer structure of the model predictive distribution.
$(ii)$ The ranking stratifies distributional structure: small $r$ emphasizes high-density regions where the model is confident, while large $r$ reflects sparser regions, highlighting uncertainty and tail behavior.

\vspace{-0.1in}
\subsection{Stage~2: Optimal Coverage Allocation}
\vspace{-0.1in}
The goal of Stage~2 is to identify the optimal coverage structure and allocation across ranks by optimizing thresholds over the distance vectors from Stage~1. This efficiency-oriented optimization produces a rank-dependent threshold vector.

For an input $X$, let $\hat Y_{j(1)},\dots,\hat Y_{j(K)}$ denote the ranked generated samples by the density-aware ranking rule from Stage~1.
For any radius $t\ge 0$, define the rank-$r$ region
\begin{equation}
\label{eq:C_r_region}
    \mathcal B_r(X;t)
    :=
    \left\{
    y\in\mathcal Y:
    \|y-\hat Y_{j(r)}\|
    \le t
    \right\}.
\end{equation}
The remaining question is how to assign a radius to each rank to maximize the prediction set efficiency while retaining coverage guarantees. To this end, consider an allocation vector
$\beta=(\beta_1,\dots,\beta_K)^\top$. For each rank $r$, let $Q_r\bigl(\{E_{1,r},\dots,E_{n_1,r}\};\beta_r\bigr)$
denote the empirical $(1-\beta_r)$-quantile of the rank-$r$ distance distribution obtained in Stage~1. For notational simplicity, we write $Q_r(\beta_r)$ when it is clear from context. Allocation vector $\beta$ induces the
rank-dependent prediction set $\bigcup_{r=1}^K \mathcal B_r\bigl(X;Q_r(\beta_r)\bigr),$ we then introduce how to find an optimal $\beta$ with efficient prediction set.

\vspace{-.1in}
\paragraph{Quantile Optimization.}
  Rather than explicitly computing the geometric union of overlapping regions, which depends heavily on complex local topologies and scales poorly, we formulate efficiency as an additive volume budget across the $K$ regions.
Specifically, we define the cost of assigning radius $Q_r(\beta_r)$ to rank $r$ as its individual volume, $Q_r(\beta_r)^d$, where $d=\dim(Y)$.
Minimizing the sum of these rank-specific volumes naturally drives the optimizer to allocate larger radii to high-density ranks (where many samples are covered at a low volume cost) and heavily penalizes expansion in sparse regions.
Coverage is enforced by
\[
\frac{1}{n_1}\sum_{i=1}^{n_1} \hat S_i(\beta) \;\ge\; 1-\alpha,
\qquad
\hat S_i(\beta)=\max_{r\in[K]} \mathbbm{1}\{E_{i,r}\le Q_r(\beta_r)\}.
\]
The resulting optimization problem is
\begin{equation}
\label{eq:optim_problem}
    \min_{\beta}\;\sum_{r=1}^K (Q_r(\beta_r))^d
    \quad \text{s.t.}\quad \frac{1}{n_1}\sum_{i=1}^{n_1} \hat S_i(\beta)\;\ge\; 1-\alpha.
\end{equation}
Intuitively, to minimize set size, the optimizer concentrates coverage in high-density regions, where small radii capture substantial probability mass. In contrast, achieving the same coverage in low-density regions requires much larger radii and is therefore inefficient.
In doing so, it effectively mirrors the oracle level set, with $\beta$ acting as a rank-dependent allocation rule that shapes the uncertainty set toward the optimal density level set.

\vspace{-.1in}
\paragraph{Equivalent MILP Reformulation.}
While the optimization \eqref{eq:optim_problem} is conceptually simple, solving it directly by grid search requires evaluating $n_1^K$ possible threshold configurations, which is computationally prohibitive. To address this, we reformulate the problem as a mixed integer linear program (MILP). This reformulation substantially enhances the scalability of our method. Theoretically, it is equivalent to the original problem and admits global optimality certificates from modern MILP solvers. Empirically, solving the MILP is often $10$--$100\times$ faster than heuristic approximation algorithms, while achieving the exact optimal solution.

For each rank $r \in \{1,\dots,K\}$, let $E_{(\ell),r}$ denote the $\ell$-th order statistic of the set $\{E_{i,r}\}_{i=1}^{n_1}$, so that $E_{(1),r} \leq E_{(2),r} \leq \cdots \leq E_{(n_1),r}$. These serve as the candidate thresholds for $Q_r(\beta_r)$. We define incremental costs
\[
    c_{r,\ell} := E_{(\ell),r}^d - E_{(\ell-1),r}^d, \qquad \ell=2,\dots,n_1,
\]
with $c_{r,1} := E_{(1),r}^d$, and coverage indicators
\[
    a_{i,r,\ell} := \mathbbm{1}\{E_{i,r} \le E_{(\ell),r}\},
    \quad i\in [n_1],\ \ell\in[n_1].
\]
Let $\tau = \lceil n_1(1-\alpha) \rceil$ denote the required number of points to cover.
We then introduce two binary variables:
$(i)$ $z_{r,\ell}=1$ if $E_{(\ell),r}$ is covered by the chose threshold and $0$ otherwise; by construction, if $z_{r,\ell}=1$ then all lower levels $\{1,\dots,\ell-1\}$ must also be $1$;
$(ii)$ $y_i=1$ if sample $i$ is covered and $0$ otherwise. The MILP problem can then be rewritten as:
\begin{subequations}\label{eq:milp}
\renewcommand{\theequation}{\theparentequation\alph{equation}}
\begin{align}
&\min_{z,y}\;\;
 \sum_{r=1}^K \sum_{\ell=1}^{n_1} c_{r,\ell}\, z_{r,\ell}
\label{eq:milp_obj}\\[-2pt]
&\text{s.t.}\;\;
 y_i \le \sum_{r=1}^K \sum_{\ell=1}^{n_1} a_{i,r,\ell}\, z_{r,\ell}\ (\forall i \in [n_1]),\;
  \sum_{i=1}^{n_1} y_i \ge \tau
\tag{\theparentequation b}\label{eq:milp_cover_quota}\\[-2pt]
& z_{r,\ell} \ge z_{r,\ell+1},\;
  z_{r,\ell'}\in\{0,1\},\;
  y_i\in\{0,1\}
  \quad
  (\forall r\in[K],\ \ell\in[n_1-1],\ \ell'\in[n_1],\ i\in[n_1]).
\tag{\theparentequation c}\label{eq:milp_prefix_binary}
\end{align}
\end{subequations}

\begin{remark}[Exact MILP Reformulation of \eqref{eq:optim_problem}]
Let $\ell_r^\ast$ denote the choosen threshold at rank $r$ with order $l^\ast$. Then by summing increments,
\[
\sum_{r=1}^K\sum_{\ell=1}^{n_1} c_{r,\ell}\ z_{r,\ell}
=\sum_{r=1}^K\sum_{\ell\le \ell_r^\ast}\!\big(E_{(\ell),r}^d - E_{(\ell-1),r}^d\big)
=\sum_{r=1}^KE_{(\ell_r^\ast),r}^d.
\]
We observe the MILP objective \eqref{eq:milp_obj} coincides with
$\sum_{r=1}^K (Q_r(\beta_r))^d$ in \eqref{eq:optim_problem}.
Constraint \eqref{eq:milp_cover_quota} encodes the coverage guarantee that at least $\tau$ points are covered.
Constraint \eqref{eq:milp_prefix_binary} enforces monotonicity across thresholds
and integrality of the decision variables. Hence the MILP \eqref{eq:milp} is an \emph{exact reformulation} of
\eqref{eq:optim_problem} with identical optimal value.
This yields Theorem~\ref{thm:milp-optimality}, with proof deferred to Appendix~\ref{sec: appdneix_proof}.
\end{remark}

\begin{theorem}[Exact MILP Reformulation and Optimality]
\label{thm:milp-optimality}
The MILP formulation \eqref{eq:milp} is equivalent to optimization \eqref{eq:optim_problem}. In particular, there exists
a one-to-one correspondence between feasible quantile vectors
$\beta=(\beta_1,\dots,\beta_K)$ in \eqref{eq:optim_problem} and feasible
binary assignments $(z,y)$ in \eqref{eq:milp}. Consequently, the optimal
solution of the MILP yields a globally optimal solution to the original
problem \eqref{eq:optim_problem}.
\end{theorem}

\paragraph{Efficient MILP Approximations.}
The MILP in \eqref{eq:milp} exactly solves the Stage~2 proxy optimization, but its worst-case runtime is exponential in the number of binary variables, which is on the order of $K n_1$. We therefore consider two scalable alternatives. The first, \texttt{ORCA}-LP, is a linear-programming (LP) relaxation of the MILP: it relaxes $(z,y)\in\{0,1\}$ to $(z,y)\in[0,1]$, yielding an LP of polynomial size with $O(Kn_1)$ variables and $O(Kn_1^2)$ nonzero constraint coefficients, and can be solved in polynomial time. The second, \texttt{ORCA}-Greedy, interprets \eqref{eq:milp} as a weighted partial set-cover problem over rank--level jumps: starting from minimal radii, it repeatedly selects the jump with the largest marginal number of newly covered exploration samples per unit proxy-volume increase, followed by a pruning step that removes redundant expansions. It has worst-case complexity $O(n_1^3K^2)$, but is substantially faster in practice; for thousands of exploration samples, it typically runs within tens of seconds in our experiments. Full algorithmic details and the greedy approximation guarantee are deferred to Appendix~\ref{app:optimization_approx}. Importantly, \texttt{ORCA}'s finite-sample validity does not depend on solving \eqref{eq:milp} exactly: Stage~2 only selects a rank-dependent radius direction, while Stage~3 applies split-conformal calibration along that direction. Thus \texttt{ORCA}-MILP, \texttt{ORCA}-LP, and \texttt{ORCA}-Greedy all preserve marginal coverage after calibration, differing only in efficiency and computational cost.

\vspace{-.1in}
\subsection{Stage-3: Conformal Calibration with Coverage Guarantees}
\vspace{-.1in}

Let $\hat\beta$ denote the solution of the Stage~2 optimization. In Stage~3, holding $\hat\beta$ fixed, we use $\mathcal{D}_{\mathrm{calib}}$ to perform the final conformal calibration.
Define the nonconformity scores
\begin{equation}
\label{eq:nonconformity_score}
    S_i := \min_{r \in [K]} \frac{E_{i,r}}{Q_r(\hat\beta_r)},
    \qquad i \in \mathcal{D}_{\text{calib}}.
\end{equation}
The score quantifies how tightly each calibration point is covered relative to
the Stage~1 threshold: $S_i<1$ indicates over-coverage and $S_i>1$ under-coverage.
We then obtain the final coverage radii $R(\alpha)=(R_1(\alpha),\dots,R_K(\alpha))^\top$, with each $R_r(\alpha) := t_\alpha \, Q_r(\hat\beta_r)$, where
\begin{equation}
\label{eq:coverage_threshold}
    t_\alpha := \min \left\{ \tau :
    \frac{1}{n_2+1} \sum_{i=1}^{n_2} \mathbbm{1}\{ S_i \leq \tau \} \;\geq\; 1-\alpha \right\}.
\end{equation}
The resulting prediction set takes the form
\begin{equation}
\label{eq:coverage_set}
     \hat{\mathcal{C}}_{\textrm{ORCA}}(X_{\text{test}}) =\bigcup_{r=1}^K{\mathcal{B}}_r(X_{\text{test}};R_r(\alpha))
    = \bigcup_{r=1}^K \Big\{ y:\ \|y - \hat{Y}_{\text{test},j(r)}\| \leq R_r(\alpha) \Big\}.
\end{equation}
\begin{theorem}[Validity of \texttt{ORCA}]
\label{thm:validity}
Suppose the test point $(X_{\text{test}},Y_{\text{test}})$ is exchangeable with the calibration data $\mathcal{D}_{\text{calib}}$.
Then the prediction set $\hat{\mathcal{C}}_{\textrm{ORCA}}(X_{\text{test}})$ satisfies
\[
    \Pr\bigl( Y_{\text{test}} \in \hat{\mathcal C}_{\mathrm{ORCA}}(X_{\text{test}}) \bigr) \;\geq\; 1 - \alpha.
\]
\end{theorem}

\noindent
The proof follows from the standard conformal prediction argument with the exchangeability assumption. The full proof is deferred to Appendix~\ref{sec: appdneix_proof}.

\vspace{-0.1in}
\subsection{Asymptotics and Level-Set Limit}
\label{subsec:asymptotics}
\vspace{-0.1in}

We now formalize the asymptotic properties of the proposed method, and all the proofs and supporting lemmas are deferred to Appendix~\ref{sec: appdneix_proof}.
\begin{proposition}[Density--Order Consistency with Averaged $m$-NN Distances]
\label{prop:density-order}
For any given $x$, let $Z_1,\ldots,Z_K \overset{\text{i.i.d.}}{\sim} p(\cdot \mid x)$ for a density
$p(\cdot \mid x)$ on a compact domain $\mathcal{Y}\subset\mathbb{R}^d$.
Assume $p(\cdot \mid x)$ is continuous and bounded away from $0$ and $\infty$.
For $z\in\mathcal{Y}$, let $\bar R_m(z) \;=\; \frac{1}{m}\sum_{j=1}^m R_j(z)$, where $R_j(z)$ is the Euclidean distance from $z$ to its $j$-th nearest neighbor among $\{Z_1,\dots,Z_K\}$.
Suppose $m \to\infty$ and $m/K\to 0$ as $K\to\infty$.
Then ordering $\{Z_j\}$ by increasing $\bar R_m(\cdot)$ coincides with ordering by decreasing $p(\cdot \mid x)$, with probability $1-o(1):$
\[
p\big(Z_{(1)}\mid x\big)\ \ge\ p\big(Z_{(2)}\mid x\big)\ \ge\ \cdots\ \ge\ p\big(Z_{(K)}\mid x\big),
\]
where $Z_{(r)}$ is the point with the $r$-th smallest $\bar R_m(\cdot)$.
\end{proposition}

\begin{theorem}[Convergence to the Optimal HDR]
\label{thm:cond-HDR}
Under the assumptions of Proposition~\ref{prop:density-order},
let $\mathcal{C}_{K,n_1,n_2}(x)$ denote the prediction set constructed from
$K$ generated samples, with $|\mathcal{D}_{\text{explore}}| = n_1$, and $|\mathcal{D}_{\text{calib}}| = n_2$.
If $K,n_1,n_2 \to \infty$, then with probability $1-o(1)$,
\[
\mu\!\left(\mathcal{C}_{K,n_1,n_2}(x)\ \triangle\ L_{\tau^\star}(x)\right)\ \to\ 0,
\]
where $L_{\tau^\star}(x) := \{ y : p(y\mid x)\ge \tau^\star(x)\}$ is the
$(1-\alpha)$ highest-density region, and $\tau^\star(x)$ is the largest cutoff satisfying
$\int_{L_{\tau^\star}(x)} p(y\mid x)\,dy = 1-\alpha$.
Hence, the Lebesgue measure of the symmetric difference between
$\mathcal{C}_{K,n_1,n_2}(x)$ and the oracle HDR $L_{\tau^\star}(x)$
converges to zero in probability.
\end{theorem}

Proposition~\ref{prop:density-order} shows that the ranking is consistent with the underlying density ordering: points in higher-density regions admit systematically smaller averaged $m$-NN radii, so sorting by $\bar R_m$ recovers the true density order.
Theorem~\ref{thm:cond-HDR} then establishes the asymptotic consequence: as $K,n_1,n_2\to\infty$, the optimization-based coverage allocation concentrates precisely on these high-density regions. Consequently, the resulting uncertainty set converges to the oracle highest-density region.

\vspace{-.1in}
\section{Experiments}
\vspace{-.1in}

In this section, we evaluate the empirical performance of \texttt{ORCA} compared with several state-of-the-art baselines on $6$ synthetic data, a semi-synthetic data, and $9$ real-world datasets. We demonstrate the superior performance of \texttt{ORCA}, especially in handling complex distributions.

\vspace{-.1in}
\paragraph{Baselines}
We compare against state-of-the-art conformal prediction methods: Conformalized Quantile Regression (CQR)~\citep{Romano:2019}, Conformal Histogram Regression (CHR)~\citep{Sesia:2021}, Distributional Conformal Prediction (DCP)~\citep{Chernozhukov:2021}, DistSplit and CDSplit~\citep{Izbicki:2020}, Probabilistic Conformal Prediction (PCP)~\citep{wang:2023}, Conformal Region Designer (CRD)~\citep{Tumu:2024}, and Regression-to-Classification Conformal Prediction (R2CCP)~\citep{Guha:2024}.
CQR and CHR output continuous sets via conditional quantiles and conditional histograms. DCP, DistSplit, and CDSplit incorporate estimated distributions/densities into conformalization. The remaining methods yield discontinuous sets: R2CCP reduces regression to classification and interpolates class scores, while CRD optimizes simple geometric templates (e.g., hyperrectangles, ellipsoids, convex hulls) to reduce volume.
We use the authors’ public implementations for CQR, CHR, R2CCP, CRD, and PCP; for DCP, DistSplit, and CDSplit we use the public implementation provided in~\cite{wang:2023}.

\vspace{-.1in}
\paragraph{Implementation Configuration}

Throughout the experiments, we set $\alpha=0.1$ and evaluate methods by marginal coverage (Cover), worst-slab conditional coverage (Cond\_Cover), and prediction set size (Eff). We set $K=20$ for experiments, except $K=50$ for the bike dataset, and set $m=\lceil\sqrt K\rceil$ for $m$-nearest-neighbor distances. For baseline modeling and conditional sample generation, we use quantile regression forests \citep{Meinshausen:2006} and mixture density networks \citep{bishop1994mixture,rothfuss2019conditional}. Full metric definitions, implementation details, data-generation procedures, and modeling details are given in Appendix~\ref{sec:Implementation_datails}.

\vspace{-.1in}
\subsection{Synthetic Data}
\vspace{-.1in}

\begin{figure}[!t]
    \centering
    \includegraphics[width=\linewidth]{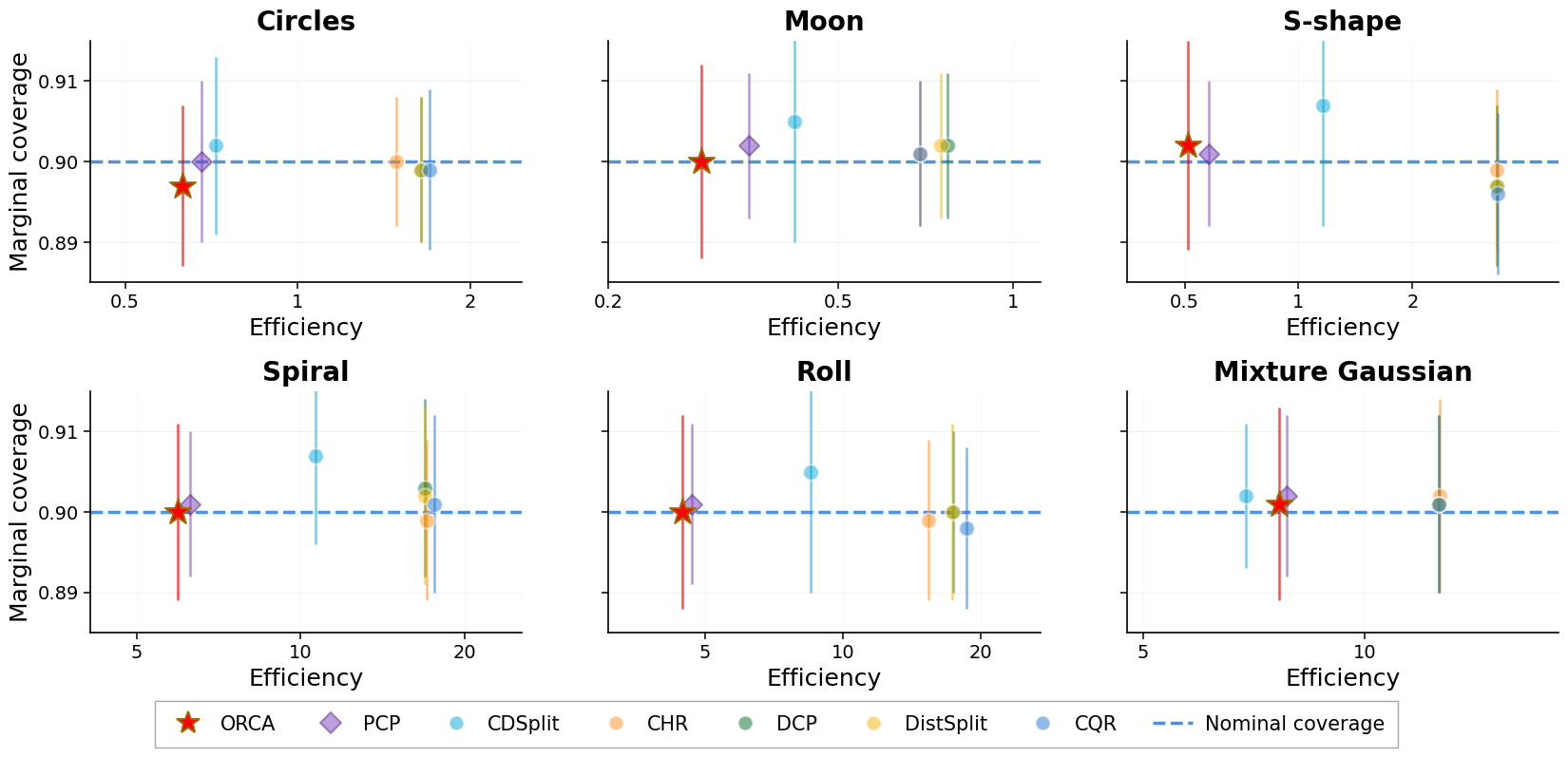}
    \caption{
    Empirical coverage and uncertainty-set efficiency on synthetic datasets. Each panel reports marginal coverage versus prediction-set efficiency on a log scale; smaller efficiency is better. The dashed horizontal line marks the nominal coverage level \(0.90\), and vertical bars show the standard error for marginal coverage. \texttt{ORCA} performs best overall.
}
    \label{fig:synthetic_results}
\end{figure}

We benchmark on six synthetic datasets (Circles, Moon, S-shape, Spiral, Roll, Mixture Gaussian) with multimodal distributions. Models are trained on $3{,}000$ points; we use $500$ for exploration, $500$ for calibration, and $1{,}000$ for testing. Results are averaged over \(30\) independent runs.
Figure~\ref{fig:synthetic_results} summarizes empirical coverage and uncertainty-set efficiency across datasets. All methods achieve near-nominal coverage, while \texttt{ORCA} attains the smallest or near-smallest uncertainty sets on nearly all datasets.
Full numerical results are provided in Appendix~\ref{sec:synthetic_data_appendix}, Table~\ref{tab:simulation}; additional visualizations are provided
in Figure~\ref{fig:1_d_synthetic_combine}.
The visualization shows that both PCP and \texttt{ORCA} produce discontinuous
uncertainty sets adapted to multimodal distributions, while \texttt{ORCA}
further improves efficiency through locally adaptive radii from optimized coverage allocation.

\begin{figure}[!t]
    \centering
\includegraphics[width=0.9\linewidth]{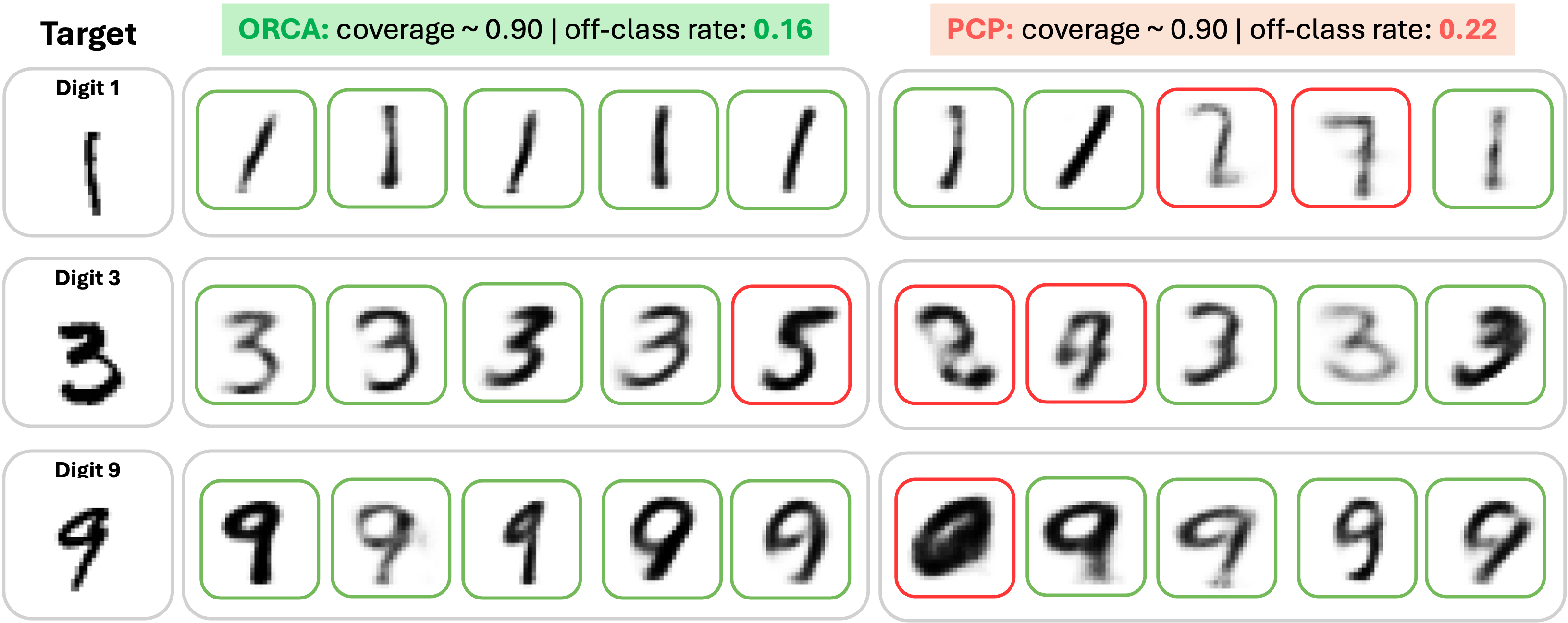}
    \caption{\textbf{MNIST digit generation coverage.} Samples from PCP and \texttt{ORCA} uncertainty sets across target digits (rows). \texttt{ORCA} yields more semantically consistent samples with target digit number and a lower off-class rate (OCR) than PCP ($0.16\pm0.02$ vs.\ $0.22\pm0.03$).}
    \label{fig:MINST_demo}
    \vspace{-0.1in}
\end{figure}

\vspace{-.1in}
\subsection{MNIST: Digit Image Generation}
\vspace{-.1in}

We evaluate \texttt{ORCA} on high-dimensional outputs using a semi-synthetic MNIST setup~\citep{deng2012mnist}. The input is the digit label $X\in\{0,\dots,9\}$ and the response $Y$ is an image of digit $X$. We train a VAE and construct uncertainty sets in the latent space.
Figure~\ref{fig:MINST_demo} shows random samples from PCP and \texttt{ORCA} uncertainty sets with $K=40$ across digits (rows). Both methods attain nominal coverage, but \texttt{ORCA}’s sampled images are visually closer to the target digit, whereas PCP more often includes less similar (and occasionally off-class) samples, indicating tighter uncertainty sets under \texttt{ORCA}. Since set size in pixel space is not directly interpretable, we report the \emph{off-class rate (OCR)}, defined as the probability that a non-\(X\) digit appears in the prediction set. Averaged over 30 runs, \texttt{ORCA} achieves a lower OCR than PCP, \(0.16\) versus \(0.22\).

\vspace{-.1in}
\subsection{Real Data}
\vspace{-.1in}

\begin{figure}[!tbh]
    \centering
    \includegraphics[width=\linewidth]{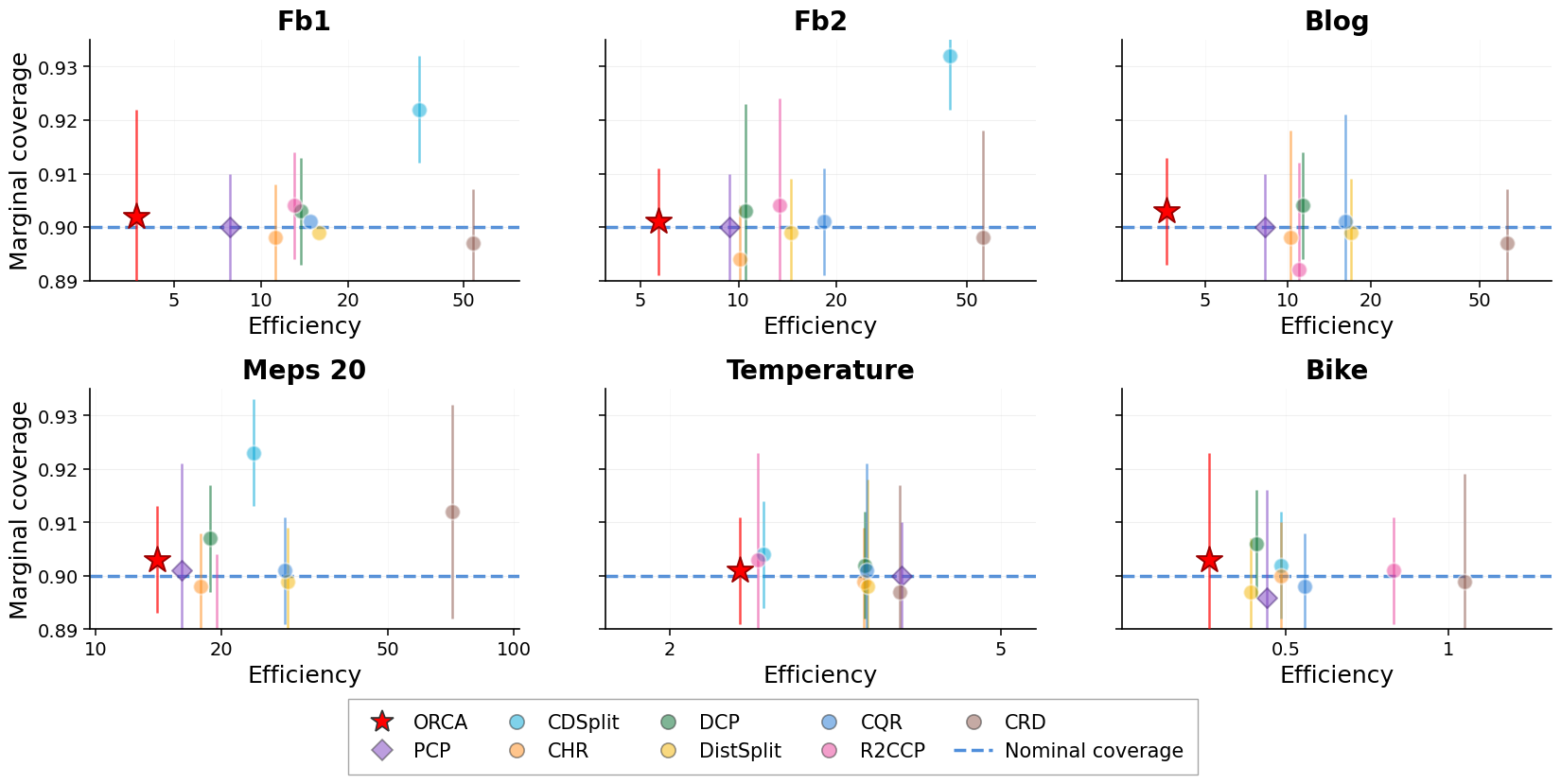}
    \caption{
    Empirical coverage and uncertainty-set efficiency on real datasets. Each panel reports marginal coverage versus prediction-set efficiency on a log scale; smaller efficiency is better. The dashed horizontal line marks the nominal coverage level \(0.90\), and vertical bars show the standard error for marginal coverage. \texttt{ORCA} performs best overall.
    }
    \label{fig:real_data_results}
\end{figure}

We evaluate \texttt{ORCA} on standard benchmarks, including Blog~\citep{blogfeedback_304}, Facebook Comment Volume (Fb1, Fb2)~\citep{facebook_comment_volume_363}, MEPS (2019--2021), Protein Tertiary Structure (Bio), and Temperature Forecasts~\citep{Cho:2020}, as well as NYC bike-sharing for 2D destination prediction. For each dataset, we allocate 500 samples for exploration, 500 for calibration, 1{,}000 for testing, and use the remaining samples for training. Consistent with \cite{Sesia:2021}, we find DCP unstable on real data; we therefore use DCP-CQR, a hybrid variant that stabilizes DCP with CQR. Figure~\ref{fig:real_data_results} reports representative real-data results, and the complete numerical results are deferred to Table~\ref{tab:real_data_summary_full} in Appendix~\ref{appendix:real_data}.

\begin{figure}[!t]
    \centering
    \captionsetup[sub]{font=footnotesize}
    \setlength{\tabcolsep}{2pt}

    \begin{subfigure}[b]{0.24\linewidth}
        \centering
        \captionsetup{justification=centering}
        {\setlength{\fboxsep}{0pt}
         \setlength{\fboxrule}{0.8pt}
         \fbox{\includegraphics[width=\linewidth]{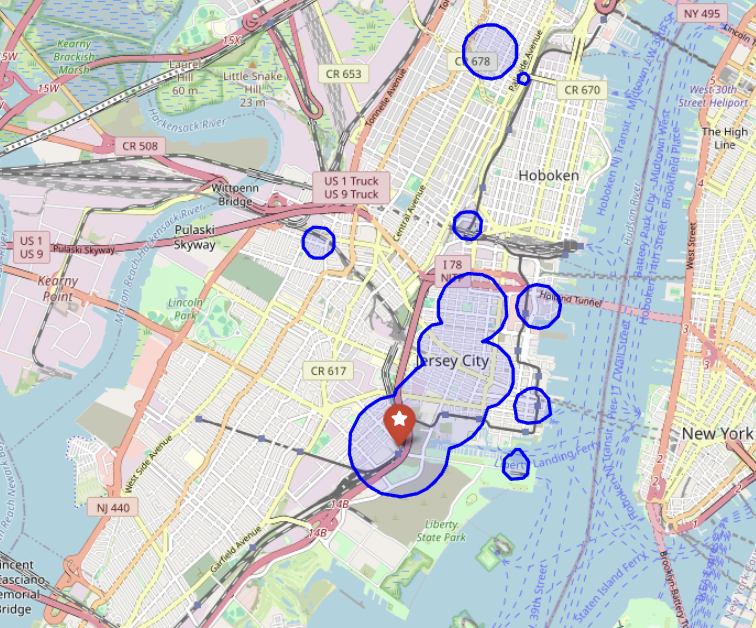}}}
        \caption{\texttt{ORCA}}
    \end{subfigure}
    \hfill
    \begin{subfigure}[b]{0.24\linewidth}
        \centering
        \captionsetup{font=footnotesize,justification=centering}
        {\setlength{\fboxsep}{0pt}
         \setlength{\fboxrule}{0.8pt}
         \fbox{\includegraphics[width=\linewidth]{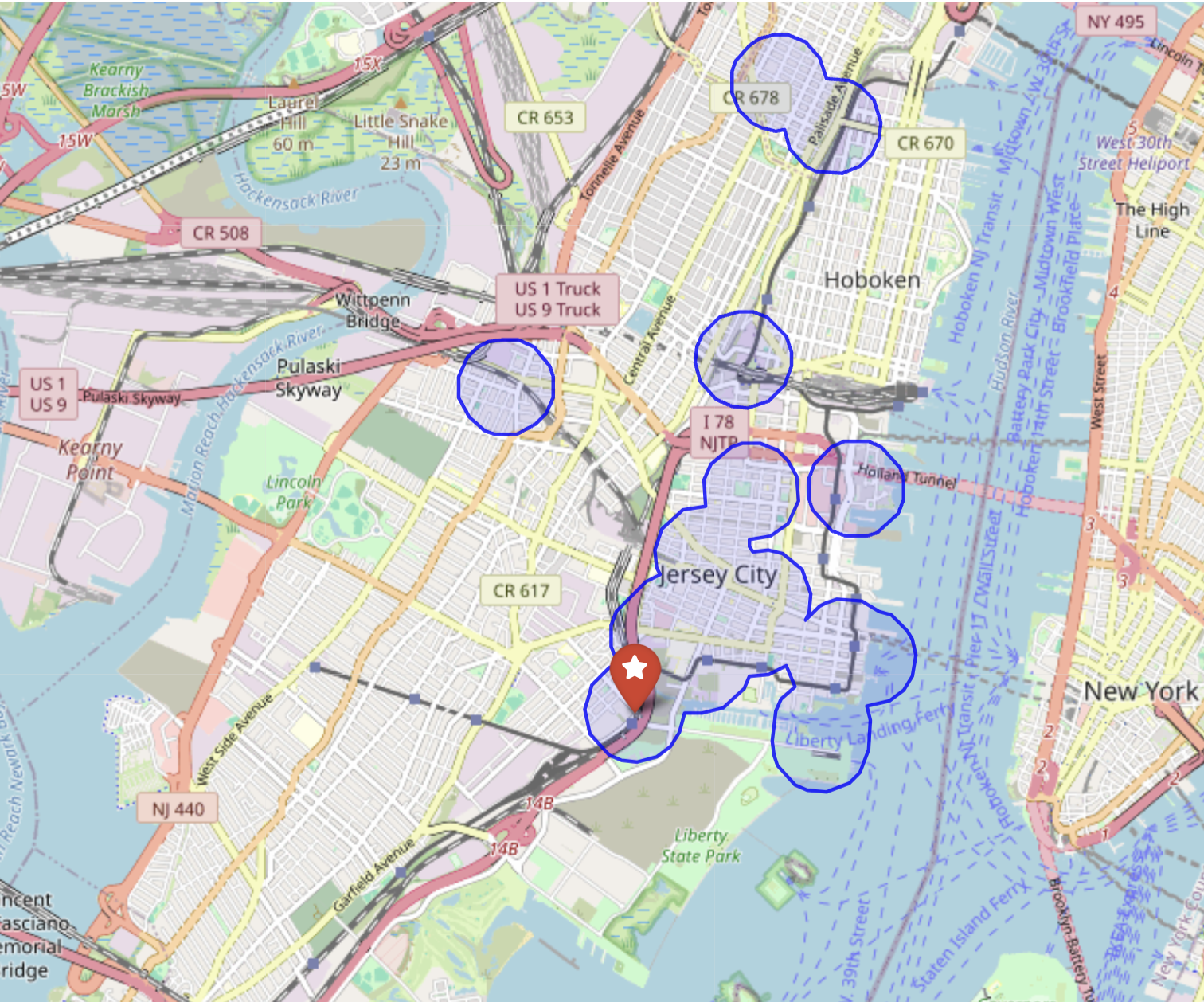}}}
        \caption{PCP}
    \end{subfigure}
    \hfill
    \begin{subfigure}[b]{0.24\linewidth}
        \centering
        \captionsetup{font=footnotesize,justification=centering}
        {\setlength{\fboxsep}{0pt}
         \setlength{\fboxrule}{0.8pt}
         \fbox{\includegraphics[width=\linewidth]{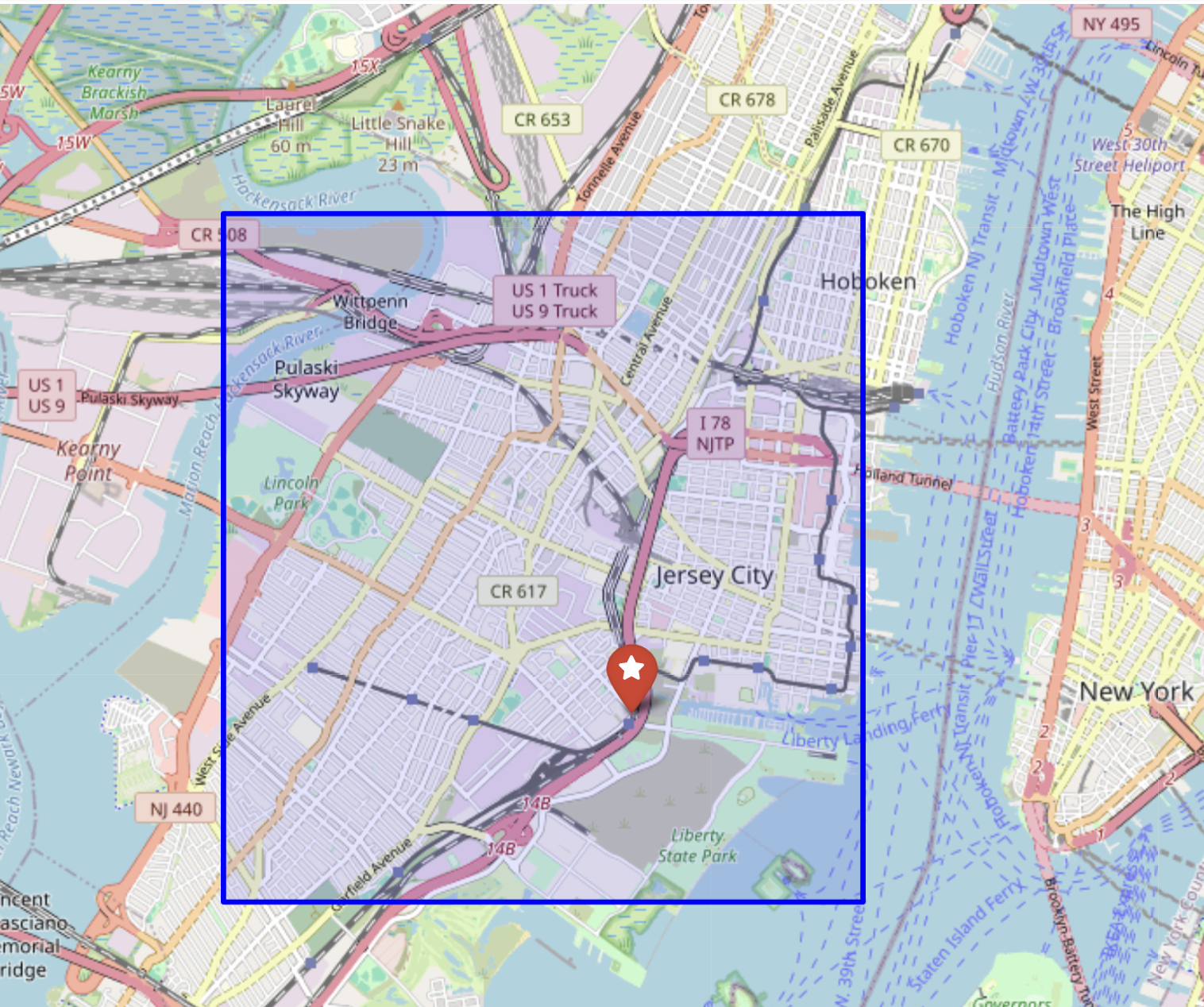}}}
        \caption{CHR}
    \end{subfigure}
    \hfill
    \begin{subfigure}[b]{0.24\linewidth}
        \centering
        \captionsetup{font=footnotesize,justification=centering}
        {\setlength{\fboxsep}{0pt}
         \setlength{\fboxrule}{0.8pt}
         \fbox{\includegraphics[width=\linewidth]{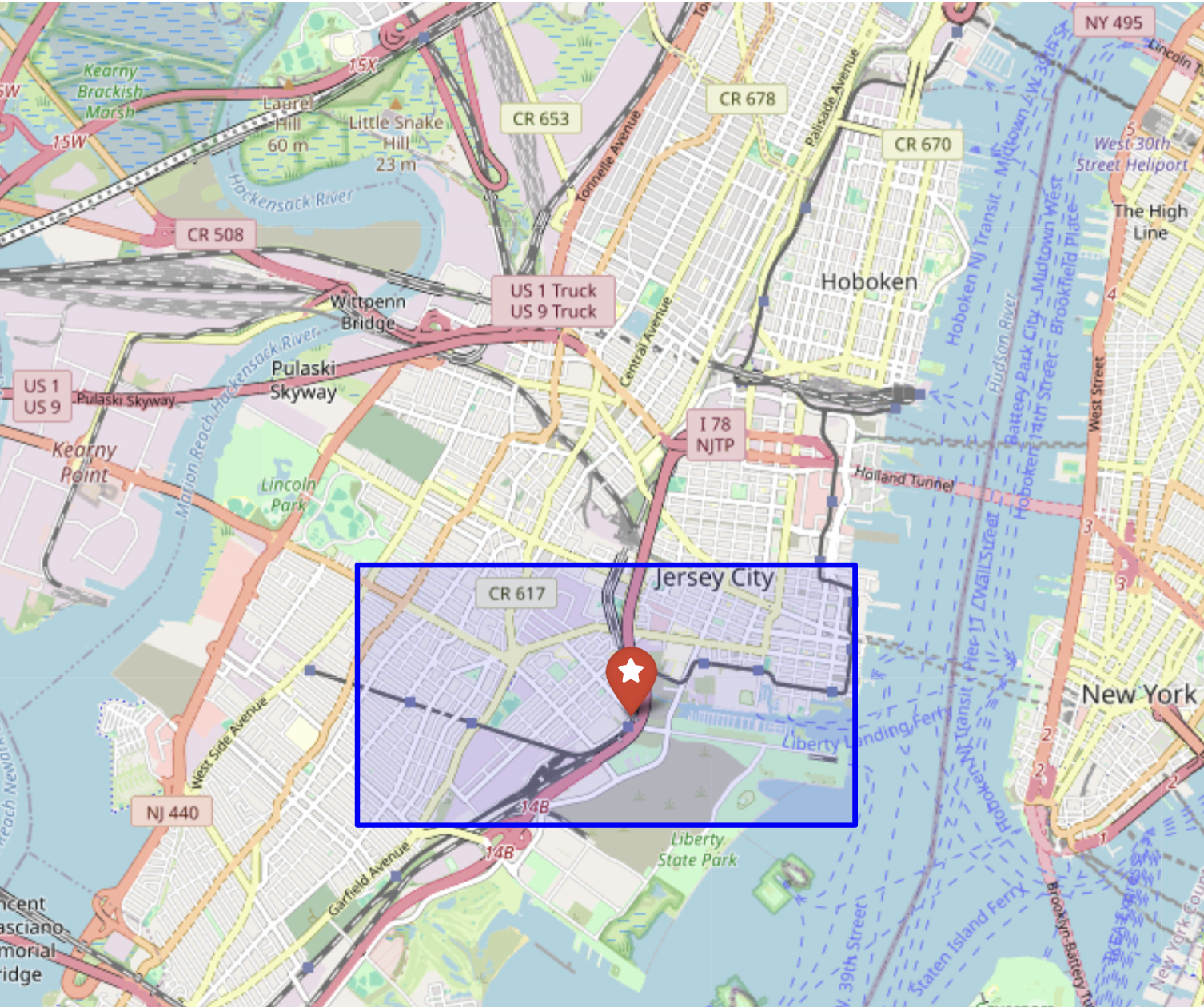}}}
        \caption{R2CCP}
    \end{subfigure}

    \caption{Comparison of prediction sets across methods with the highest efficiency on the Bike dataset. The colored region indicates the set, and the red marker denotes the target location.}
    \label{fig:bike_comparison}
    \vspace{-0.15in}
\end{figure}

Figure~\ref{fig:real_data_results} shows that all methods achieve coverage close to the nominal level, while \texttt{ORCA} consistently yields smaller prediction sets. The gains are especially pronounced on complex datasets such as Blog, where the conditional distribution \(P(Y\mid X)\) is highly structured and benefits from rank-dependent coverage allocation. By assigning larger radii to high-density regions and shrinking radii in low-density regions, \texttt{ORCA} avoids unnecessary expansion into sparse areas. In contrast, methods based on a single global scale or simple geometric templates tend to over-cover low-density regions in order to maintain marginal coverage, leading to larger sets.
Figure~\ref{fig:bike_comparison} further illustrates this behavior on NYC bike destination prediction. Both PCP and \texttt{ORCA} produce discontinuous prediction regions as unions of local balls, whereas CHR and R2CCP return continuous rectangular regions. Such discontinuity is desirable in this setting because destinations naturally cluster around transit hubs, with low-density regions in between. Compared with PCP, \texttt{ORCA} uses optimized coverage allocation across ranks to suppress unnecessary expansion around low-density generated samples, producing tighter and more targeted uncertainty sets without sacrificing coverage.
\vspace{-.1in}
\subsection{Sensitivity analysis}
\label{sec:sensitivity}
\vspace{-.1in}

To evaluate the practical deployment of \texttt{ORCA}, we conduct a comprehensive set of sensitivity analyses. Specifically, we examine $(i)$ robustness to generative model misspecification, $(ii)$ the influence of the $m$-NN ranking parameter, which controls the local density proxy used to order generated samples; $(iii)$ the runtime--efficiency trade-off among the exact MILP formulation, our efficient approximation methods, and baseline approaches; and $(iv)$ metric choice in high-dimensional output spaces. For the last point, we further evaluate \texttt{ORCA} on MNIST using several metrics, including learned latent metrics, motivated by the fact that standard $\ell_p$ distances are often poorly suited for high-dimensional image spaces. Overall, these analyses show that \texttt{ORCA} maintains consistent performance gains under moderate parameter choices, remains robust to reasonable levels of model misspecification, and admits scalable approximations for computationally demanding settings. Full experimental details and results are provided in Appendix~\ref{appendix:sensitivity_analysis}.

\vspace{-.1in}
\section{Discussion and Future Work}
\vspace{-.1in}

\texttt{ORCA} combines generative modeling with rank-adaptive coverage allocation to construct flexible prediction regions while preserving finite-sample marginal coverage. The proposed MILP exactly solves the Stage~2 allocation problem, and the LP and greedy variants offer scalable alternatives with little empirical loss in efficiency. Several directions remain for future work. First, in high-dimensional output spaces, the choice of metric is important, since Euclidean distance may not reflect semantic similarity; learned metrics may further improve performance. Second, the Stage~2 objective uses an additive volume cost over rank-dependent regions; a more direct objective that accounts for the union size of overlapping regions could further improve efficiency.

\section*{Acknowledgments}
This work was supported by the National Science Foundation under Grant No.~CAIG-2425888.

\bibliography{ref}
\bibliographystyle{unsrt}
\newpage
\appendix

\section{Average $m$-Nearest Neighbor Distance}
\label{append:avg_m_nn_dist}

The average $m$-nearest neighbor distance for a generated sample $\hat{Y}_{i,k}$ can be computed as follows:
\begin{enumerate}
\item (Optional) Sample a sequence of reference samples $\{\tilde{Y}_{i,l}\}_{l=1}^M \sim \hat{p}(\cdot|X_i)$ for ranking estimation. Otherwise, set $\tilde{Y}_{i,l} = \hat{Y}_{i,l}$ for $l=1,\dots,K$ and $M=K$.
    \item
    Compute the pairwise distances of generated samples among reference samples:
    \[
        D_{i}(k, l) \coloneqq \|\hat{Y}_{i,k} - \tilde{Y}_{i,l}\|,~\text{for}~k \in \{1,\dots,K\}, \ l \in \{1,\dots,M\}.
    \]
\item Sort the distances to $\hat{Y}_{i,k}$ such that:
    \[
        D_{i}(k, (1)) \le \dots \le D_{i}(k, (M)),
    \]
    where $D_{i}(k, (l))$ denotes the $l$-th smallest distance to $\hat{Y}_{i,k}$.  \item Calculate average $m$-nearest neighbor distance:
    \[
        \bar{D}_{i,k} = \frac{1}{m} \sum_{l=1}^m D_{i}(k, (l)).
    \]
\end{enumerate}

\section{Comparison with PCP}
\label{sec: compare_with_pcp}
Probabilistic Conformal Prediction (PCP) \citep{wang:2023} first studied CP for generative models and utilized the information of random samples. In contrast to our method, which considers all the distances between each random sample and the true response, PCP defines its non-conformity score as the minimum distance between random samples and the true response:
\begin{equation}
\label{eq:PCP_non_conformity_score}
    E_i = \min_{k=1,\dots,K} \|Y_i - \hat{Y}_{i,k}\|.
\end{equation}
for each calibration data $(X_i,Y_i)$.
For a new test data $(X_{n+1}, Y_{n+1})$, the prediction set of PCP is constructed as the union of $K$ regions:
\begin{equation}
\label{eq:PCP_coverage_set}
    \hat{\mathcal{C}}(X_{n+1}) = \cup_{k=1}^K \hat{\mathcal{C}}_k,
\end{equation}
where each region $\hat{\mathcal{C}}_k$ is defined by
\[
    \hat{\mathcal{C}}_k = \left \{y: \|y - \hat{Y}_{n+1,k}\| \leq Q(E_{1:n} \cup \{\infty\}; \alpha) \right \}.
\]
Here $Q(\cdot; \alpha)$ is the $1-\alpha$ quantile of the empirical distribution of the non-conformity scores $E_1, \dots, E_n$.

As noted, PCP adopts a scalar non-conformity score, which assigns a uniform radius to all prediction regions $\hat{\mathcal C}_k$.
This uniform treatment of random samples limits adaptability, as ideally, high-density regions should have larger radius, while low-density regions require smaller ones.
In contrast, \texttt{ORCA} introduces a vectorized non-conformity score that captures the full complexity of the prediction error distribution.
By ranking samples based on their empirical density, our method enables individualized optimization of the ball radius for each sample. As demonstrated in Figure \ref{fig:bike_comparison}, the ball radius for low-density regions nearly diminishes, highlighting the flexibility of our approach. Another limitation of PCP arises when
$K$ is small. In such cases, the empirical distribution of minimum distances may become heavy-tailed, requiring a large quantile (radius) to maintain coverage, leading to an overly conservative prediction set $\hat{\mathcal{C}}(X_{n+1})$.
In contrast, the proposed \texttt{ORCA} overcomes this by introducing an individualized quantile selection, relaxing the constraint for attaining the $1-\alpha$ quantile uniformly across all samples at each rank and extensively reducing the impact of the distribution's tail.
\section{Proof}
\label{sec: appdneix_proof}
\subsection{Proof of Theorem \ref{thm:milp-optimality}}
\begin{proof}
For each rank $r$, let $\{E_{(1),r}\le\cdots\le E_{(n_1),r}\}$ denote the order statistics of $\{E_{i,r}\}_{i=1}^{n_1}$, and define incremental costs
\[
c_{r,1}=E_{(1),r}^d,\qquad
c_{r,\ell}=E_{(\ell),r}^d-E_{(\ell-1),r}^d \ \ (\ell\ge 2).
\]

\noindent\textbf{(i) From quantile problem to MILP.}
Let $\beta$ be feasible for \eqref{eq:optim_problem}. For each $r$, let $\ell_r$ satisfy
$Q_r(\beta_r)=E_{(\ell_r),r}$ and set $z_{r,\ell}:=\mathbbm{1}\{\ell\le \ell_r\}$.
By the definition of $y_i$, we have
\[
y_i:=\mathbbm{1}\!\left\{\sum_{r=1}^K\sum_{\ell=1}^{n_1} a_{i,r,\ell}\,z_{r,\ell}\ge 1\right\}=\max_{r\in[K]}\mathbbm{1}\{E_{i,r}\le Q_r(\beta_r)\}.
\]
Then:

\emph{(a) Monotonicity \eqref{eq:milp_prefix_binary}.}
By choice of $\ell_r$ and construction $z_{r,\ell}$, $z_{r,\ell}\ge z_{r,\ell+1}$ for all $\ell$, and $z_{r,\ell}\in\{0,1\}$.
Also $y_i\in\{0,1\}$ by definition.

\emph{(b) Coverage quota \eqref{eq:milp_cover_quota}.} We have
$y_i \le \sum_{r,\ell} a_{i,r,\ell}z_{r,\ell}$ and
\[
\sum_{i=1}^{n_1} y_i
= \sum_{i=1}^{n_1} \max_{r\in[K]}\mathbbm{1}\{E_{i,r}\le Q_r(\beta_r)\}
\;\ge\; \tau,
\]
by the feasibility of $\beta$ \eqref{eq:optim_problem}.

\emph{(c) Objective \eqref{eq:milp_obj}.}
By telescoping of increments,
\[
\sum_{r=1}^K\sum_{\ell=1}^{n_1} c_{r,\ell}\,z_{r,\ell}
=\sum_{r=1}^K\sum_{\ell=1}^{\ell_r} \bigl(E_{(\ell),r}^d - E_{(\ell-1),r}^d\bigr)
=\sum_{r=1}^K E_{(\ell_r),r}^d
=\sum_{r=1}^K \bigl(Q_r(\beta_r)\bigr)^d.
\]
Thus $(z,y)$ is feasible for \eqref{eq:milp} and preserves the objective.

\medskip
\noindent\textbf{(ii) From MILP to quantile problem.}
Let $(z,y)$ be feasible for \eqref{eq:milp}. For each $r$, set
$\ell_r^\star:=\max\{\ell: z_{r,\ell}=1\}$ (define $\ell_r^\star=0$ if all $z_{r,\ell}=0$),
$Q_r^\star:=E_{(\ell_r^\star),r}$, and pick $\beta_r^\star$ with $Q_r(\beta_r^\star)=Q_r^\star$.
Then
\[
\max_{r}\mathbbm{1}\{E_{i,r}\le Q_r^\star\}
= \mathbbm{1}\!\left\{\sum_{r,\ell} a_{i,r,\ell}z_{r,\ell}\ge 1\right\}\ \ge\ y_i,
\]
so $\sum_i \max_r \mathbbm{1}\{E_{i,r}\le Q_r^\star\}\ge \sum_i y_i\ge \tau$, i.e., $\beta^\star$ is feasible for \eqref{eq:optim_problem}. Moreover,
\[
\sum_{r=1}^K \bigl(Q_r(\beta_r^\star)\bigr)^d
=\sum_{r=1}^K E_{(\ell_r^\star),r}^d
=\sum_{r=1}^K\sum_{\ell=1}^{n_1} c_{r,\ell}\,z_{r,\ell},
\]
so the objective values coincide.

\noindent\textbf{(iii) Equivalence.}
The two optimization problems are feasible and objective-preserving in both directions, establishing an one-to-one correspondence between solutions of \eqref{eq:optim_problem} and \eqref{eq:milp}. Hence the MILP is an exact reformulation of the original problem, with identical feasible sets and optimal values.
\end{proof}

\subsection{Proof of Theorem \ref{thm:validity}}
\begin{proof}
Let $\mathcal{D}_{\text{calib}}=\{(X_i,Y_i)\}_{i=1}^{n_2}$ be the calibration set and $(X_{\text{test}},Y_{\text{test}})$ the test point that is exchangeable with $\mathcal{D}_{\text{calib}}$.
Condition on the exploration split $\D_{\mathrm{explore}}$ and the optimized threshold $Q_r(\hat\beta_r)$ for each $r$ from Stage~2, which are fixed before calibration.
Define the nonconformity scores
\[
S_i=\min_{r\in[K]}\frac{E_{i,r}}{Q_r(\hat\beta_r)}, \quad i=1,\dots,n_2,
\qquad
S_{\text{test}}=\min_{r\in[K]}\frac{E_{\text{test},r}}{Q_r(\hat\beta_r)}.
\]
The calibration factor $t_\alpha$ is computed:
\[
t_\alpha=\min\Bigl\{\tau\in \mathbb{R}:\frac{1}{n_2+1}\sum_{i=1}^{n_2}\mathbbm{1}\{S_i\le \tau\}\ge 1-\alpha\Bigr\}.
\]
By exchangeable assumption, the scores $\{S_1,\dots,S_{n_2},S_{\text{test}}\}$ are exchangeable, and the rank of each $S_i$ is uniformly distributed.
Therefore
\[
\Pr\;\bigl(S_{\text{test}}\le t_\alpha\bigr)\;\ge\;1-\alpha.
\]
Finally, we have the following equivalence
\begin{equation}
    \begin{aligned}
    S_{\text{test}} \le t_\alpha
    &\iff \min_{r\in[K]} \frac{E_{\text{test},r}}{Q_r(\hat\beta_r)} \le t_\alpha
    \\
    &\iff \exists\, r\in[K]\ \text{s.t.}\ E_{\text{test},r}\le t_\alpha\, Q_r(\hat\beta_r)
    \\
    &\iff \exists\, r\in[K]\ \text{s.t.}\ \|Y_{\text{test}}-\hat Y_{\text{test},\,j(r)}\|\le t_\alpha\, Q_r(\hat\beta_r)
    \\
    &\iff \exists\, r\in[K]\ \text{s.t.}\ \|Y_{\text{test}}-\hat Y_{\text{test},\,j(r)}\|\le R_r(\alpha)
    \quad (\text{since }R_r(\alpha):=t_\alpha Q_r(\hat\beta_r))
    \\
    &\iff Y_{\text{test}}\in \bigcup_{r=1}^K \Bigl\{y:\ \|y-\hat Y_{\text{test},\,j(r)}\|\le R_r(\alpha)\Bigr\}
    \\
    &\iff Y_{\text{test}}\in \hat{\mathcal{C}}_{\textrm{ORCA}}(X_{\text{test}}).
    \end{aligned}
\end{equation}
Thus
\[
\Pr\;\bigl(Y_{\text{test}}\in \hat{\mathcal{C}}_{\textrm{ORCA}}(X_{\text{test}})\bigr)=\Pr\;\bigl(S_{\text{test}}\leq t_\alpha\bigl)\;\ge\;1-\alpha,
\]
which completes the proof.
\end{proof}

\subsection{Proof of Proposition \ref{prop:density-order}}
\begin{proof}
Let $v_d$ denote the volume of the unit ball in $\mathbb{R}^d$.
A classical asymptotic result in the nearest--neighbor estimation literature states that if $m=m_K\to\infty$ with $m/K\to0$ as $K\to\infty$, then the $j$-th nearest--neighbor radius satisfies
\[
R_j(z) \;=\; \Bigl(\tfrac{j}{K\,p(z\mid x)\,v_d}\Bigr)^{1/d}\,(1+o(1)),
\]
with probability $1-o(1)$, uniformly over $z\in\Y$ and $1\le j\le m$ \citep{Loftsgaarden:1965,Zhao:2022}.

Averaging over $j=1,\dots,m$ gives
\[
\bar R_m(z)
=\frac{1}{m}\sum_{j=1}^m R_j(z)
= \bigl(v_d\,p(z\mid x)\bigr)^{-1/d}K^{-1/d}
\cdot \frac{1}{m}\sum_{j=1}^m j^{1/d}\,(1+o(1)).
\]

Since $t\mapsto t^{1/d}$ is increasing, the integral comparison
\[
\int_0^m t^{1/d}\,dt
\;\le\; \sum_{j=1}^m j^{1/d}
\;\le\; 1+\int_1^m t^{1/d}\,dt
\]
implies
\[
\sum_{j=1}^m j^{1/d}
= \frac{d}{d+1}\,m^{1+1/d} + O(1).
\]
Therefore,
\[
\frac{1}{m}\sum_{j=1}^m j^{1/d}
= \frac{d}{d+1}\,m^{1/d}\,(1+o(1)),
\]
and hence
\[
\bar R_m(z)
= (v_d)^{-1/d}\,\frac{d}{d+1}\,p(z\mid x)^{-1/d}\,K^{-1/d}\,m^{1/d}\,(1+o(1)).
\]

Thus $\bar R_m(z)$ is, up to a constant independent of $z$, proportional to $p(z\mid x)^{-1/d}$.
In particular, if $p(z\mid x) > p(z'\mid x)$, then $\bar R_m(z) < \bar R_m(z')$ with probability $1-o(1)$,
so the averaged nearest-neighbor distance ranking preserves the oracle density ordering.
Applying this comparison uniformly to all sample points $\{Z_k\}_{k=1}^K$ shows that the ordering induced by
$\bar R_m(\cdot)$ coincides with that of $p(\cdot\mid x)$, which completes the proof.
\end{proof}

\subsection{Proof of Theorem \ref{thm:cond-HDR}}
\begin{proof}[Proof of Theorem~\ref{thm:cond-HDR}]
Fix $x$ and write $p(\cdot)=p(\cdot\mid x)$. Assume $\mathcal Y$ compact, $p$ continuous with
$0<c_-\le p\le c_+<\infty$, and $\mu\{p=\tau^\star(x)\}=0$. For simplicity, here we assume the samples in exploration and calibration splits are i.i.d., and the generated samples are i.i.d. from $p(\cdot)$.

\paragraph{Step 1: Empirical Optimization to Population Optimization.}
For rank $r\in[K]$, let $F_{r,K}$ be the c.d.f.\ of $E_{i,r}$ and $\hat F_{r,K,n_1}$ its empirical c.d.f.
By DKW inequality and a union bound argument,
\[
\max_{1\le r\le K}\ \sup_{\beta\in[\varepsilon,1-\varepsilon]}
\bigl|Q_r(\beta)-q_r(\beta)\bigr|=O_{\mathbb P}\!\Big(\sqrt{\tfrac{\log K}{n_1}}\Big)\to0,
\]
assuming the positivity of the density of $F_{r,K}$. Define the empirical and
population programs
\[
\min_{\beta\in[\varepsilon,1-\varepsilon]^K}\ J_{K,n_1}(\beta):=\sum_{r=1}^K Q_r(\beta_r)^d
\quad\text{s.t.}\quad
C_{K,n_1}(\beta):=\tfrac1{n_1}\sum_{i=1}^{n_1}\max_r\mathbf 1\{E_{i,r}\le Q_r(\beta_r)\}\ge 1-\alpha,
\]
\[
\min_{\beta\in[\varepsilon,1-\varepsilon]^K}\ J_{K}(\beta):=\sum_{r=1}^K q_r(\beta_r)^d
\quad\text{s.t.}\quad
C_{K}(\beta):=\mathbb E\big[\max_r\mathbf 1\{E_r\le q_r(\beta_r)\}\big]\ge 1-\alpha.
\]

For the constraint part, we can directly apply the Glivenko–Cantelli theorem and derive
\[
\sup_\beta|C_{K,n_1}(\beta)-C_K(\beta)|\to 0.
\]
For the objective, we can use the uniform consistency of quantile function, so that
\begin{equation}
    \sup_\beta|J_{K,n_1}(\beta)-J_K(\beta)|\to0
\end{equation}
Hence the empirical optimizer $\hat\beta_{K,n_1}$ is asymptotically optimal for the population program
(argmin consistency). Let
\[
\hat{\mathcal C}_{K,n_1}(x)=\bigcup_{r=1}^K \mathcal B\!\left(Z_{(r)},\,Q_r(\hat\beta_{K,n_1,r})\right),\qquad
\mathcal C_K^*(x)=\bigcup_{r=1}^K \mathcal B\!\left(Z_{(r)},\,q_r(\beta_{K,r}^*)\right),
\]
where $\beta_K^*$ is a population optimizer.

\paragraph{Step 2: Population cutoff and convergence to HDR.}
By Proposition~\ref{prop:density-order}, with probability $1-o(1)$ the order $r\mapsto Z_{(r)}$ matches decreasing
$p$. Consider the population program with objective $\sum_r q_r(\beta_r)^d$ and coverage constraint
$C_K(\beta)\ge 1-\alpha$. The objective is coordinatewise increasing in the radii
$\{q_r(\beta_r)\}$, while, for small increments of radius, the mass gain per volume around center
$Z_{(r)}$ is proportional to $p(Z_{(r)})$ (first-orde expansion).
The threshold $\tau_K$ will be in the form of
such that
\[
q_r(\beta_{K,r}^*)>0\ \Rightarrow\ p(Z_{(r)})\ge \tau_K,\qquad
q_r(\beta_{K,r}^*)=0\ \text{ otherwise.}
\]
The radii of the ball will be vanish with order $O(K^{-1/d})$
Finally, the coverage constraint $C_K(\beta_K^*)=1-\alpha$ forces $\tau_K\to\tau^\star(x)$, the unique level with
$\int_{L_{\tau^\star}} p=1-\alpha$, hence
\[
\mu\!\left(\mathcal C_K^*(x)\ \triangle\ L_{\tau^\star}(x)\right)\ \to\ 0.
\]

\paragraph{Step 3: Calibration.}
Split-conformal scales radii by $t_\alpha$. Since the pre-calibrated set already attains asymptotic
coverage $1-\alpha$ and the score distribution concetrates at $1$, $t_\alpha\to 1$ in probability, so
\[
\mu\!\left(\mathcal C_{K,n_1,n_2}(x)\ \triangle\ L_{\tau^\star}(x)\right)\ \xrightarrow{\ \mathbb P\ }\ 0.
\]
\end{proof}

\section{Optimization Approximations}
\label{app:optimization_approx}

This section provides the algorithmic details for the scalable Stage~2 optimization variants used in our experiments. Recall that the exact MILP in \eqref{eq:milp} optimizes over prefix variables $z_{r,\ell}$ and coverage indicators $y_i$. For each rank $r\in[K]$, let
\[
E_{(1),r}\le \cdots \le E_{(n_1),r}
\]
denote the order statistics of $\{E_{i,r}\}_{i=1}^{n_1}$, and set $E_{(0),r}=0$. We use the same incremental costs and coverage indicators as in the main text:
\[
c_{r,\ell}:=E_{(\ell),r}^d-E_{(\ell-1),r}^d,
\qquad
a_{i,r,\ell}:=\mathbbm{1}\{E_{i,r}\le E_{(\ell),r}\}.
\]
We also write $a_{i,r,0}=0$ and let
\[
\tau=\lceil n_1(1-\alpha)\rceil
\]
be the required number of exploration samples to cover.

For any integer level vector $\ell=(\ell_1,\dots,\ell_K)\in\{0,\dots,n_1\}^K$, define the induced rank-dependent radii
\[
q_r(\ell):=E_{(\ell_r),r},
\qquad r\in[K],
\]
and the corresponding covered set
\[
U(\ell)
:=
\left\{
i\in[n_1]:
\exists r\in[K]\ \text{such that}\ E_{i,r}\le E_{(\ell_r),r}
\right\}.
\]
The Stage~2 feasibility condition is $|U(\ell)|\ge \tau$, and the proxy objective is
\[
\mathrm{Cost}(\ell)
:=
\sum_{r=1}^K q_r(\ell)^d
=
\sum_{r=1}^K E_{(\ell_r),r}^d
=
\sum_{r=1}^K\sum_{h=1}^{\ell_r}c_{r,h}.
\]
The exact MILP searches over all feasible level vectors. The two approximation methods below return feasible integer level vectors $\hat\ell$, and therefore rank-dependent radii
\[
\hat q_r=E_{(\hat\ell_r),r},\qquad r\in[K].
\]
These radii are then passed to Stage~3, where the final split-conformal rescaling is applied.

\paragraph{\texttt{ORCA}-LP: LP relaxation and threshold rounding.}

The first approximation relaxes the binary constraints in \eqref{eq:milp} to obtain a linear program:
\begin{subequations}
\label{eq:lp_relaxation}
\begin{align}
\min_{z,y}\quad
& \sum_{r=1}^K\sum_{\ell=1}^{n_1}c_{r,\ell}z_{r,\ell} \\
\mathrm{s.t.}\quad
& y_i \le \sum_{r=1}^K\sum_{\ell=1}^{n_1}a_{i,r,\ell}z_{r,\ell},
\qquad i\in[n_1],\\
& \sum_{i=1}^{n_1}y_i\ge \tau,\\
& z_{r,\ell}\ge z_{r,\ell+1},
\qquad r\in[K],\ \ell\in[n_1-1],\\
& 0\le z_{r,\ell}\le 1,
\qquad r\in[K],\ \ell\in[n_1],\\
& 0\le y_i\le 1,
\qquad i\in[n_1].
\end{align}
\end{subequations}
The LP relaxation provides a lower bound on the exact MILP objective. To recover a feasible integer solution, we round the fractional prefix variables by thresholding. Since the LP preserves the prefix constraints $z_{r,\ell}\ge z_{r,\ell+1}$, thresholding the fractional solution at any level $\gamma$ produces a valid prefix vector. We scan over all distinct fractional values and choose the cheapest thresholded solution that covers at least $\tau$ exploration samples.

\begin{algorithm}[H]
\caption{\textbf{\texttt{ORCA}-LP: LP relaxation with prefix rounding}}
\label{alg:orca_lp}
\small
\begin{algorithmic}[1]
\STATE \textbf{Input}: distance matrix $\{E_{i,r}\}$; costs $\{c_{r,\ell}\}$; indicators $\{a_{i,r,\ell}\}$; coverage quota $\tau$
\STATE Solve the LP relaxation \eqref{eq:lp_relaxation} and obtain fractional solution $(\tilde z,\tilde y)$
\STATE Let $\Gamma=\{0,1\}\cup\{\tilde z_{r,\ell}:r\in[K],\ell\in[n_1]\}$, sorted in decreasing order
\FOR{$\gamma\in\Gamma$}
    \STATE Set $z^{(\gamma)}_{r,\ell}=\mathbbm{1}\{\tilde z_{r,\ell}\ge \gamma\}$ for all $r,\ell$
    \STATE Define $\ell_r^{(\gamma)}=\max\{\ell:z^{(\gamma)}_{r,\ell}=1\}$, with $\ell_r^{(\gamma)}=0$ if no such $\ell$ exists
    \STATE Compute the covered set
    \[
    U^{(\gamma)}
    =
    \left\{
    i\in[n_1]:
    \exists r\in[K]\ \text{s.t.}\ a_{i,r,\ell_r^{(\gamma)}}=1
    \right\}
    \]
    \IF{$|U^{(\gamma)}|\ge \tau$}
        \STATE Set $\hat\ell_r=\ell_r^{(\gamma)}$ for all $r\in[K]$
        \STATE \textbf{break}
    \ENDIF
\ENDFOR
\STATE \textbf{Output}: rank-dependent radii $\hat q_r=E_{(\hat\ell_r),r}$, $r\in[K]$
\end{algorithmic}
\end{algorithm}

Because $\gamma=0$ selects all prefix variables, the rounding procedure always returns a feasible solution. In practice, this LP-guided rounding often achieves an objective close to the exact MILP while being substantially faster for moderate or large $K$ and $n_1$.

\paragraph{\texttt{ORCA}-Greedy: jump-based coverage expansion.}

The second approximation directly exploits the partial-cover structure of \eqref{eq:milp}. Instead of increasing one prefix level at a time, \texttt{ORCA}-Greedy considers \emph{jumps}: from a current level $\ell_r$, it may move rank $r$ directly to any higher level $h>\ell_r$. This allows the method to skip uninformative intermediate thresholds and substantially reduces runtime.

At a current level vector $\ell$, define the current covered set $U(\ell)$. A jump $(r,h)$ with $h>\ell_r$ changes only the rank-$r$ level from $\ell_r$ to $h$. Its marginal coverage gain is
\[
\Delta_{\mathrm{cov}}(r,h;\ell)
:=
\left|
\left\{
i\notin U(\ell):
a_{i,r,h}=1
\right\}
\right|,
\]
and its marginal cost is
\[
\Delta_{\mathrm{cost}}(r,h;\ell)
:=
E_{(h),r}^d-E_{(\ell_r),r}^d
=
\sum_{s=\ell_r+1}^{h}c_{r,s}.
\]
The greedy rule selects the jump with the largest gain-to-cost ratio:
\[
(r^\star,h^\star)
\in
\arg\max_{r\in[K],\,h>\ell_r}
\frac{\Delta_{\mathrm{cov}}(r,h;\ell)}
{\Delta_{\mathrm{cost}}(r,h;\ell)}.
\]
After reaching the coverage quota, we apply a backward pruning step that lowers redundant rank levels whenever doing so preserves feasibility.

\begin{algorithm}[H]
\caption{\textbf{\texttt{ORCA}-Greedy: jump-based partial cover}}
\label{alg:orca_greedy}
\small
\begin{algorithmic}[1]
\STATE \textbf{Input}: distance matrix $\{E_{i,r}\}$; costs $\{c_{r,\ell}\}$; indicators $\{a_{i,r,\ell}\}$; coverage quota $\tau$
\STATE Initialize $\ell_r=0$ for all $r\in[K]$
\STATE Initialize $U=\emptyset$
\WHILE{$|U|<\tau$}
    \FOR{$r\in[K]$ and $h\in\{\ell_r+1,\dots,n_1\}$}
        \STATE Compute
        \[
        \Delta_{\mathrm{cov}}(r,h)
        =
        \left|
        \left\{
        i\notin U:
        a_{i,r,h}=1
        \right\}
        \right|
        \]
        \STATE Compute
        \[
        \Delta_{\mathrm{cost}}(r,h)
        =
        E_{(h),r}^d-E_{(\ell_r),r}^d
        \]
    \ENDFOR
    \STATE Select
    \[
    (r^\star,h^\star)
    \in
    \arg\max_{r,h:\Delta_{\mathrm{cov}}(r,h)>0}
    \frac{\Delta_{\mathrm{cov}}(r,h)}
    {\Delta_{\mathrm{cost}}(r,h)}
    \]
    \STATE Update $\ell_{r^\star}\leftarrow h^\star$
    \STATE Update
    \[
    U\leftarrow
    \left\{
    i\in[n_1]:
    \exists r\in[K]\ \text{s.t.}\ a_{i,r,\ell_r}=1
    \right\}
    \]
\ENDWHILE
\STATE \textbf{Backward pruning:}
\STATE Set $\mathrm{changed}\leftarrow \mathrm{true}$
\WHILE{$\mathrm{changed}$}
    \STATE Set $\mathrm{changed}\leftarrow \mathrm{false}$
    \FOR{$r\in[K]$}
        \STATE Find the smallest $h\in\{0,\dots,\ell_r\}$ such that replacing $\ell_r$ by $h$ still gives $|U(\ell)|\ge\tau$
        \IF{$h<\ell_r$}
            \STATE Set $\ell_r\leftarrow h$
            \STATE Set $\mathrm{changed}\leftarrow \mathrm{true}$
        \ENDIF
    \ENDFOR
\ENDWHILE
\STATE \textbf{Output}: rank-dependent radii $\hat q_r=E_{(\ell_r),r}$, $r\in[K]$
\end{algorithmic}
\end{algorithm}

The jump-based greedy method is especially fast because it avoids solving a global integer program. Its objective is not guaranteed to match the MILP optimum, but it returns a feasible Stage~2 radius direction whenever the loop terminates. Moreover, the backward pruning step can only reduce the proxy objective while maintaining the required exploration coverage. As with the exact MILP and LP relaxation, the final finite-sample validity of \texttt{ORCA}-Greedy is ensured by the independent Stage~3 conformal calibration; the approximation affects only the efficiency of the resulting prediction set, not its marginal coverage guarantee.

\paragraph{Approximation guarantee for the greedy variant.}

The greedy procedure can be viewed as a cost-sensitive partial set-cover algorithm over the finite ground set $[n_1]$. Each jump increases a monotone coverage function
\[
F(\ell):=|U(\ell)|
\]
and incurs a nonnegative separable cost. Let $\mathrm{OPT}_{\mathrm{MILP}}$ denote the optimal objective value of the exact MILP, and let $\mathrm{Cost}_{\mathrm{Greedy}}$ denote the cost of the greedy solution before final conformal calibration. By the standard greedy analysis for weighted partial cover,
\[
\mathrm{Cost}_{\mathrm{Greedy}}
\le
(1+\log \tau)\,\mathrm{OPT}_{\mathrm{MILP}}.
\]
The pruning step can only decrease this cost. This guarantee is for the Stage~2 proxy objective; the Stage~3 conformal rescaling is then applied in the same way as for the exact MILP solution.

\section{Implementation Details}
\label{sec:Implementation_datails}

All conformal calibration and evaluation experiments were run locally on a Mac machine with an Apple M4 chip, a 10-core CPU, and 16GB unified memory, without requiring GPU acceleration. Neural-network-based conditional density models were trained on an NVIDIA RTX 5070 Ti GPU with 16GB memory using Python 3.7. Runtime measurements for the Stage~2 optimization procedures, including the exact MILP and approximation variants, are reported in Appendix~\ref{appendix:sensitivity_analysis}. In particular, Table~\ref{tab:runtime-eff} summarizes the wall-clock runtime and efficiency tradeoff for different choices of $K$.

\subsection{Model Training and Hyperparameter Settings}
For conditional distribution estimation in the $1$-dimensional case, we implemented quantile regression forests (QRF) \citep{Meinshausen:2006} with $100$ trees, a minimum of $50$ samples per node split, and a quantile grid ranging from $1\%$ to $99\%$. The trained model serves as the base model for CQR, CHR, DCP, DistSplit, CRD, PCP, and \texttt{ORCA}. The implementation is based on the publicly available repository.\footnote{\url{https://github.com/msesia/chr}}
For methods requiring explicit density estimation, such as CDsplit, we employed a mixture density network (MDN) \citep{bishop1994mixture} with $10$ Gaussian components, dropout rate of $0.2$, hidden layer size of $(100, 100)$, and $500$ training epochs. The implementation is based on the Python package \texttt{cde} \citep{rothfuss2019conditional}.
For R2CCP, we used the default configuration from the author's public repository and trained for $500$ epochs. In the multivariate setting, we applied Bonferroni correction to construct coverage intervals for CHR, CQR, DCP, DistSplit, CDsplit, and R2CCP, using the same base models as in the $1$-dimensional case. For CRD, PCP, and \texttt{ORCA}, we used MDN to model multivariate responses.

\subsection{Real Datasets Details}
We evaluate the performance of \texttt{ORCA} and baseline methods on a diverse collection of real-world datasets spanning a range of application domains. These datasets vary in input dimensionality, response dimension, and sample size, and include both univariate and multivariate prediction tasks.
A summary of the datasets used in our experiments is provided in Table~\ref{tab:dataset_summary}.

\subsection{Evaluation Metrics}
\label{sec:evaluation_metrics}

Throughout the experiments, we set $\alpha=0.1$ and evaluate each method using three metrics: marginal coverage, worst-slab conditional coverage, and prediction set efficiency. For a test set $\{(X_i,Y_i)\}_{i=1}^{n_{\mathrm{test}}}$ and prediction sets $\hat{\mathcal C}(X_i)$, the empirical marginal coverage is defined as
\[
\mathrm{Cover}
:=
\frac{1}{n_{\mathrm{test}}}
\sum_{i=1}^{n_{\mathrm{test}}}
\mathbbm{1}\{Y_i\in \hat{\mathcal C}(X_i)\}.
\]
This measures whether the prediction sets attain the target marginal coverage level $1-\alpha$.

To evaluate conditional coverage behavior, we also report worst-slab conditional coverage. This diagnostic scans geometric slices of the covariate space and reports the lowest empirical coverage over sufficiently large slabs. Specifically, for a direction $v\in\mathbb R^{d_x}$ and scalars $a<b$, define the slab
\[
S_{v,a,b}:=\{x\in\mathbb R^{d_x}: a\le v^\top x\le b\}.
\]
Given a test set $\{(X_i,Y_i)\}_{i=1}^{n_{\mathrm{test}}}$, let
\[
\mathcal I_{v,a,b}
:=
\{i: X_i\in S_{v,a,b}\}.
\]
For a minimum mass threshold $\delta\in(0,1]$ and a finite set of candidate directions $\mathcal V$, we define the empirical worst-slab conditional coverage as
\[
\mathrm{Cond\_Cover}
:=
\min_{v\in\mathcal V}
\;
\min_{a<b:\, |\mathcal I_{v,a,b}|/n_{\mathrm{test}}\ge \delta}
\frac{1}{|\mathcal I_{v,a,b}|}
\sum_{i\in\mathcal I_{v,a,b}}
\mathbbm{1}\{Y_i\in \hat{\mathcal C}(X_i)\}.
\]
Here, $\delta$ prevents the diagnostic from selecting slabs containing too few test points. In practice, $\mathcal V$ is chosen as a finite collection of random projection directions. This metric captures the worst empirical coverage over localized geometric regions of the covariate space and serves as a diagnostic for conditional coverage heterogeneity.

Finally, we measure prediction set efficiency by the average set size,
\[
\mathrm{Eff}
:=
\frac{1}{n_{\mathrm{test}}}
\sum_{i=1}^{n_{\mathrm{test}}}
\mu\bigl(\hat{\mathcal C}(X_i)\bigr),
\]
where $\mu(\cdot)$ denotes Lebesgue measure for continuous outputs and cardinality for discrete outputs. Smaller values of $\mathrm{Eff}$ indicate more informative prediction sets, provided that coverage is maintained.

\begin{table}[!t]
\centering
\caption{Summary of real-world datasets.}
\label{tab:dataset_summary}
\begin{tabular}{lccc}
\toprule
\textbf{Dataset} & \textbf{\# Features} & \textbf{Response Dimension} & \textbf{\# Samples} \\
\midrule
Blog Feedback (Blog) \citep{Buza:2014}          & $280$   & $1$ & $52{,}397$ \\
Facebook Comment Volume (Fb1) \citep{Singh:2015} & $53$    & $1$ & $40{,}948$ \\
Facebook Comment Volume (Fb2)                  & $53$    & $1$ & $81{,}311$ \\
MEPS (Year 19)\tablefootnote{\url{https://meps.ahrq.gov/mepsweb/}} & $139$   & $1$ & $15{,}785$ \\
MEPS (Year 20)                                  & $139$   & $1$ & $17{,}541$ \\
MEPS (Year 21)                                  & $139$   & $1$ & $15{,}656$ \\
Protein Structure (Bio) \citep{Rana:2013}        & $9$     & $1$ & $45{,}730$ \\
Temperature \citep{temperature:2020}            & $21$    & $1$ & $7{,}590$ \\
Bike\tablefootnote{\url{https://citibikenyc.com/system-data}} & $11$    & $2$ & $735{,}502$ \\
\bottomrule
\end{tabular}
\end{table}

\begin{figure*}[!t]
     \centering
    \includegraphics[width=\linewidth]{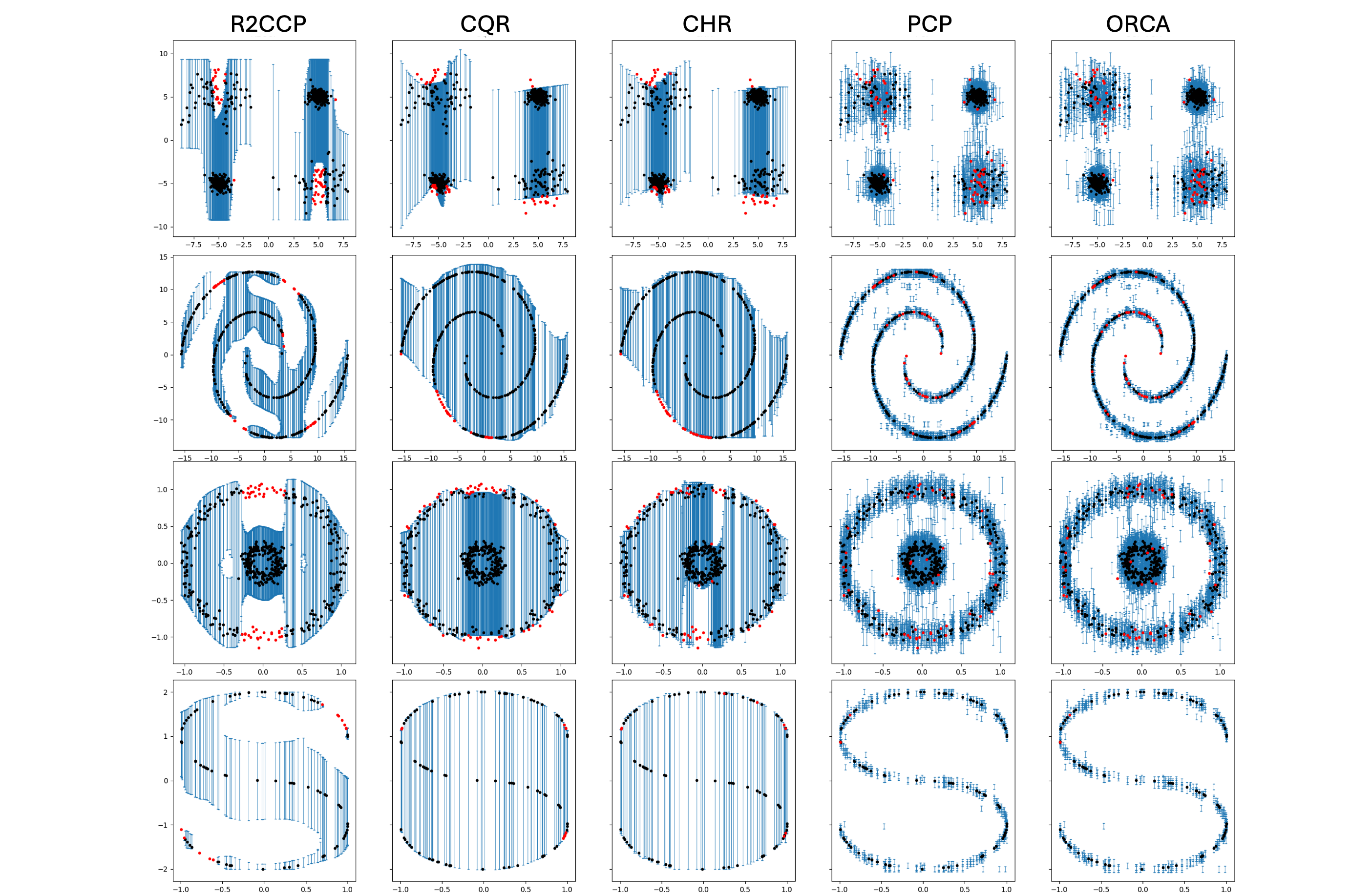}
     \caption{Comparison of prediction set ($\alpha=0.1$) generated by different methods on $1$-dimensional synthetic data: S-shape, Circles, Spirals, and Unbalanced clusters (from top to bottom row). Blues lines: the predictive intervals of each method;
Black dots: test points that are covered  by the predictive sets; Reds dots: test points not covered. We plot the figure with $100$, $500$, $400$, and $500$ random samples for each data, respectively.}
     \label{fig:1_d_synthetic_combine}
\end{figure*}

\begin{figure}[!t]
    \centering
    \includegraphics[width=\linewidth]{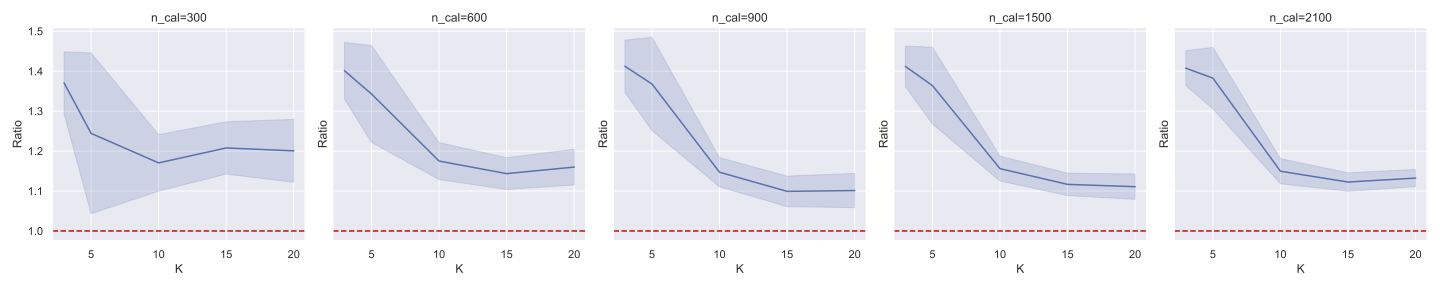}
    \caption{Ratio of PCP prediction set length over \texttt{ORCA} prediction set length under different $K=3,5,10,15,20$ and number of calibration data for S-shape data. The blue curve represents the mean, and the shaded area indicates the standard error of the ratio. We observe that the ratio remains consistently above 1, demonstrating that \texttt{ORCA} achieves better efficiency.}
    \label{fig:S-shape_K_comparison}
\end{figure}

\section{Experiments Results}
\label{appendix:experiment}
In this section, we present the additional experimental results for synthetic data, and real data.
\subsection{Synthetic Data Results}
\label{sec:synthetic_data_appendix}
We report the full synthetic-data results in Table~\ref{tab:simulation}, including marginal coverage, worst-slab conditional coverage, and efficiency, with means and standard errors computed over 30 runs.
We also provide visual comparisons of prediction sets produced by different methods on the S-shape, Circles, Spirals, and Unbalanced Clusters datasets in Figure~\ref{fig:1_d_synthetic_combine}. The results show that \texttt{ORCA} consistently produces adaptive and discontinuous prediction sets, leading to improved efficiency.
We further study the effect of $K$ on efficiency in Figure~\ref{fig:S-shape_K_comparison}, which reports the ratio of PCP prediction-set length to \texttt{ORCA} prediction-set length on the S-shape dataset for $K\in\{3,5,10,15,20\}$ and varying calibration sample sizes. The curves denote the mean ratio, and the shaded bands indicate standard errors over 100 runs. The ratio remains consistently above 1 across all settings, confirming that \texttt{ORCA} yields more efficient prediction sets than PCP. The ratio gradually decreases as $K$ increases, suggesting that the relative efficiency gain of \texttt{ORCA} becomes smaller for larger $K$.

\renewcommand{\arraystretch}{1}
\setlength{\tabcolsep}{3.5pt}
\begin{table*}[!tbh]
\vspace{-4pt}
\centering
\caption{Summary of synthetic data results.}
\vspace{-6pt}
\resizebox{\linewidth}{!}{\begin{threeparttable}
\begin{tabular}{
  C{1.9cm} C{1.9cm} C{1.9cm} C{1.9cm} C{1.9cm} C{1.9cm}
  C{1.9cm} C{1.9cm} C{1.9cm} C{1.9cm} C{1.9cm}
}
\toprule[1pt]
Dataset & Metric &
CQR & CHR & DCP & DistSplit & CDSplit &
CRD & R2CCP & PCP & \textbf{\texttt{ORCA}} \\
\midrule
\multirow{3}{*}{Circles}
& \small Cover & \small $0.90\pm0.01$ & \small $0.90\pm0.01$ & \small $0.90\pm0.01$ & \small $0.90\pm0.01$ & \small $0.90\pm0.01$ & \small $0.90\pm0.01$ & \small $0.90\pm0.01$ & \small $0.90\pm0.01$ & \cellcolor[HTML]{C0C0C0}\small $0.90\pm0.01$ \\
& \small Cond\_Cover & \small $0.85\pm0.04$ & \small $0.84\pm0.05$ & \small $0.86\pm0.03$ & \small $0.86\pm0.03$ & \small $0.87\pm0.03$ & \small $0.80\pm0.03$ & \small $0.86\pm0.01$ & \small $0.85\pm0.03$ & \cellcolor[HTML]{C0C0C0}\small $0.86\pm0.03$ \\
& \small Eff & \small $1.70\pm0.01$ & \small $1.49\pm0.06$ & \small $1.64\pm0.04$ & \small $1.64\pm0.05$ & \small $0.72\pm0.04$ & \small $1.61\pm0.01$ & \small $2.77\pm0.04$ & \small $0.68\pm0.31$ & \cellcolor[HTML]{C0C0C0}\small $\mathbf{0.63\pm0.24}$ \\
\midrule
\multirow{3}{*}{Moon}
& \small Cover & \small $0.90\pm0.01$ & \small $0.90\pm0.01$ & \small $0.90\pm0.01$ & \small $0.90\pm0.01$ & \small $0.92\pm0.01$ & \small $0.90\pm0.01$ & \small $0.90\pm0.01$ & \small $0.90\pm0.01$ & \cellcolor[HTML]{C0C0C0}\small $0.90\pm0.02$ \\
& \small Cond\_Cover & \small $0.85\pm0.05$ & \small $0.87\pm0.08$ & \small $0.87\pm0.06$ & \small $0.85\pm0.06$ & \small $0.87\pm0.05$ & \small $0.78\pm0.03$ & \small $0.86\pm0.02$ & \small $0.85\pm0.03$ & \cellcolor[HTML]{C0C0C0}\small $0.85\pm0.04$ \\
& \small Eff & \small $0.69\pm0.05$ & \small $0.69\pm0.08$ & \small $0.77\pm0.06$ & \small $0.75\pm0.06$ & \small $0.42\pm0.03$ & \small $0.33\pm0.03$ & \small $1.84\pm0.11$ & \small $0.35\pm0.04$ & \cellcolor[HTML]{C0C0C0}\small $\mathbf{0.29\pm0.04}$ \\
\midrule
\multirow{3}{*}{S-shape}
& \small Cover & \small $0.90\pm0.01$ & \small $0.90\pm0.01$ & \small $0.90\pm0.01$ & \small $0.90\pm0.01$ & \small $0.91\pm0.02$ & \small $0.90\pm0.01$ & \small $0.90\pm0.01$ & \small $0.90\pm0.02$ & \cellcolor[HTML]{C0C0C0}\small $0.90\pm0.01$ \\
& \small Cond\_Cover & \small $0.53\pm0.04$ & \small $0.80\pm0.07$ & \small $0.80\pm0.06$ & \small $0.80\pm0.07$ & \small $0.88\pm0.02$ & \small $0.81\pm0.04$ & \small $0.85\pm0.02$ & \small $0.84\pm0.03$ & \cellcolor[HTML]{C0C0C0}\small $0.83\pm0.04$ \\
& \small Eff & \small $3.36\pm0.02$ & \small $3.34\pm0.02$ & \small $3.34\pm0.02$ & \small $3.34\pm0.02$ & \small $1.16\pm0.11$ & \small $2.63\pm0.25$ & \small $6.94\pm0.47$ & \small $0.58\pm0.01$ & \cellcolor[HTML]{C0C0C0}\small $\mathbf{0.51\pm0.05}$ \\
\midrule
\multirow{3}{*}{Spiral}
& \small Cover & \small $0.90\pm0.01$ & \small $0.90\pm0.01$ & \small $0.90\pm0.01$ & \small $0.90\pm0.01$ & \small $0.91\pm0.01$ & \small $0.90\pm0.01$ & \small $0.90\pm0.01$ & \small $0.90\pm0.01$ & \cellcolor[HTML]{C0C0C0}\small $0.90\pm0.01$ \\
& \small Cond\_Cover & \small $0.75\pm0.06$ & \small $0.81\pm0.06$ & \small $0.82\pm0.07$ & \small $0.80\pm0.06$ & \small $0.87\pm0.04$ & \small $0.81\pm0.05$ & \small $0.87\pm0.02$ & \small $0.85\pm0.04$ & \cellcolor[HTML]{C0C0C0}\small $0.85\pm0.05$ \\
& \small Eff & \small $17.59\pm0.40$ & \small $17.02\pm0.58$ & \small $16.94\pm0.55$ & \small $16.92\pm0.53$ & \small $10.67\pm0.98$ & \small $12.62\pm1.27$ & \small $45.26\pm2.83$ & \small $6.27\pm0.68$ & \cellcolor[HTML]{C0C0C0}\small $\mathbf{5.96\pm0.57}$ \\
\midrule
\multirow{3}{*}{Roll}
& \small Cover & \small $0.90\pm0.01$ & \small $0.90\pm0.01$ & \small $0.90\pm0.01$ & \small $0.90\pm0.01$ & \small $0.91\pm0.01$ & \small $0.90\pm0.01$ & \small $0.90\pm0.01$ & \small $0.90\pm0.01$ & \cellcolor[HTML]{C0C0C0}\small $0.90\pm0.02$ \\
& \small Cond\_Cover & \small $0.65\pm0.07$ & \small $0.81\pm0.05$ & \small $0.77\pm0.05$ & \small $0.76\pm0.06$ & \small $0.84\pm0.04$ & \small $0.71\pm0.05$ & \small $0.86\pm0.03$ & \small $0.84\pm0.04$ & \cellcolor[HTML]{C0C0C0}\small $0.83\pm0.04$ \\
& \small Eff & \small $18.59\pm0.17$ & \small $15.33\pm0.42$ & \small $17.39\pm0.49$ & \small $17.33\pm0.47$ & \small $8.50\pm0.48$ & \small $14.64\pm0.92$ & \small $36.51\pm2.56$ & \small $4.67\pm0.57$ & \cellcolor[HTML]{C0C0C0}\small $\mathbf{4.46\pm0.58}$ \\
\midrule
\multirow{3}{*}{\shortstack{Mixture\\Gaussian}}
& \small Cover & \small $0.90\pm0.01$ & \small $0.90\pm0.01$ & \small $0.90\pm0.01$ & \small $0.90\pm0.01$ & \cellcolor[HTML]{C0C0C0}\small $0.91\pm0.01$ & \small $0.90\pm0.01$ & \small $0.90\pm0.02$ & \small $0.90\pm0.01$ & \small $0.90\pm0.02$ \\
& \small Cond\_Cover & \small $0.89\pm0.04$ & \small $0.89\pm0.04$ & \small $0.89\pm0.04$ & \small $0.89\pm0.05$ & \cellcolor[HTML]{C0C0C0}\small $0.87\pm0.03$ & \small $0.82\pm0.03$ & \small $0.85\pm0.03$ & \small $0.85\pm0.03$ & \small $0.86\pm0.03$ \\
& \small Eff & \small $12.61\pm0.11$ & \small $12.66\pm0.10$ & \small $12.62\pm0.11$ & \small $12.61\pm0.10$ & \cellcolor[HTML]{C0C0C0}\small $\mathbf{6.89\pm0.19}$ & \small $13.64\pm0.43$ & \small $14.49\pm0.80$ & \small $7.84\pm0.53$ & \small $7.66\pm0.58$ \\
\bottomrule[1pt]
\end{tabular}
\begin{tablenotes}
\item[1] \footnotesize Metrics include marginal coverage (Cover), worst-slab conditional coverage (Cond\_Cover), and prediction set efficiency (Eff). Values report mean $\pm$ standard error over $30$ runs. The lowest efficiency per dataset is bolded and highlighted along with corresponding coverage metrics.
\end{tablenotes}
\end{threeparttable}}
\label{tab:simulation}
\end{table*}

\subsection{Real Data Results}
We report the full real datasets results in Table~\ref{tab:real_data_summary_full}, including marginal coverage, worst-slab conditional coverage, and efficiency, with means and standard errors computed over 30 runs.

\label{appendix:real_data}
\renewcommand{\arraystretch}{1}
\setlength{\tabcolsep}{3.5pt}
\begin{table*}[!tbh]
\vspace{-4pt}
\centering
\caption{Summary of real data results.}
\vspace{-6pt}
\resizebox{\linewidth}{!}{\begin{threeparttable}
\begin{tabular}{
  C{1.9cm} C{1.9cm} C{1.9cm} C{1.9cm} C{1.9cm} C{1.9cm}
  C{1.9cm} C{1.9cm} C{1.9cm} C{1.9cm} C{1.9cm}
}
\toprule[1pt]
Dataset & Metric &
CQR & CHR & DCP-CQR & DistSplit & CDSplit &
CRD & R2CCP & PCP & \textbf{\texttt{ORCA}} \\
\midrule
\multirow{3}{*}{Fb1}
& \small Cover & \small $0.90\pm0.00$ & \small $0.90\pm0.01$ & \small $0.90\pm0.01$ & \small $0.90\pm0.00$ & \small $0.92\pm0.01$ & \small $0.90\pm0.01$ & \small $0.90\pm0.01$ & \small $0.90\pm0.01$ & \cellcolor[HTML]{C0C0C0}\small $0.90\pm0.02$ \\
& \small Cond\_Cover & \small $0.87\pm0.02$ & \small $0.86\pm0.05$ & \small $0.88\pm0.00$ & \small $0.89\pm0.03$ & \small $0.87\pm0.07$ & \small $0.83\pm0.03$ & \small $0.76\pm0.04$ & \small $0.82\pm0.03$ & \cellcolor[HTML]{C0C0C0}\small $0.80\pm0.05$ \\
& \small Eff & \small $14.78\pm2.31$ & \small $11.23\pm1.34$ & \small $13.73\pm1.55$ & \small $15.88\pm2.07$ & \small $35.09\pm2.69$ & \small $54.07\pm40.50$ & \small $12.97\pm0.82$ & \small $7.79\pm0.28$ & \cellcolor[HTML]{C0C0C0}\small $\mathbf{3.70\pm1.75}$ \\
\midrule
\multirow{3}{*}{Fb2}
& \small Cover & \small $0.90\pm0.01$ & \small $0.89\pm0.01$ & \small $0.90\pm0.02$ & \small $0.90\pm0.01$ & \small $0.93\pm0.01$ & \small $0.90\pm0.02$ & \small $0.90\pm0.02$ & \small $0.90\pm0.01$ & \cellcolor[HTML]{C0C0C0}\small $0.90\pm0.01$ \\
& \small Cond\_Cover & \small $0.88\pm0.02$ & \small $0.84\pm0.04$ & \small $0.89\pm0.05$ & \small $0.89\pm0.01$ & \small $0.87\pm0.08$ & \small $0.84\pm0.03$ & \small $0.73\pm0.05$ & \small $0.82\pm0.03$ & \cellcolor[HTML]{C0C0C0}\small $0.80\pm0.04$ \\
& \small Eff & \small $18.31\pm1.57$ & \small $10.12\pm1.04$ & \small $10.54\pm1.12$ & \small $14.48\pm1.43$ & \small $44.41\pm1.61$ & \small $56.26\pm2.64$ & \small $13.39\pm1.65$ & \small $9.35\pm0.21$ & \cellcolor[HTML]{C0C0C0}\small $\mathbf{5.68\pm1.62}$ \\
\midrule
\multirow{3}{*}{Blog}
& \small Cover & \small $0.90\pm0.02$ & \small $0.90\pm0.02$ & \small $0.90\pm0.01$ & \small $0.90\pm0.01$ & \small $0.96\pm0.01$ & \small $0.90\pm0.01$ & \small $0.89\pm0.02$ & \small $0.90\pm0.01$ & \cellcolor[HTML]{C0C0C0}\small $0.90\pm0.01$ \\
& \small Cond\_Cover & \small $0.87\pm0.02$ & \small $0.89\pm0.03$ & \small $0.90\pm0.04$ & \small $0.88\pm0.03$ & \small $0.91\pm0.02$ & \small $0.81\pm0.03$ & \small $0.74\pm0.06$ & \small $0.81\pm0.04$ & \cellcolor[HTML]{C0C0C0}\small $0.82\pm0.05$ \\
& \small Eff & \small $16.18\pm2.61$ & \small $10.26\pm1.77$ & \small $11.37\pm1.76$ & \small $16.93\pm2.13$ & \small $39.95\pm3.30$ & \small $63.07\pm37.05$ & \small $10.95\pm1.82$ & \small $8.27\pm0.21$ & \cellcolor[HTML]{C0C0C0}\small $\mathbf{3.61\pm1.27}$ \\
\midrule
\multirow{3}{*}{Bio}
& \small Cover & \small $0.90\pm0.01$ & \cellcolor[HTML]{C0C0C0}\small $0.90\pm0.01$ & \small $0.90\pm0.01$ & \small $0.90\pm0.01$ & \small $0.90\pm0.01$ & \small $0.90\pm0.01$ & \small $0.89\pm0.01$ & \small $0.90\pm0.01$ & \small $0.90\pm0.01$ \\
& \small Cond\_Cover & \small $0.90\pm0.03$ & \cellcolor[HTML]{C0C0C0}\small $0.90\pm0.04$ & \small $0.89\pm0.03$ & \small $0.90\pm0.04$ & \small $0.86\pm0.03$ & \small $0.84\pm0.04$ & \small $0.83\pm0.03$ & \small $0.85\pm0.02$ & \small $0.82\pm0.03$ \\
& \small Eff & \small $13.09\pm0.18$ & \cellcolor[HTML]{C0C0C0}\small $\mathbf{9.44\pm0.33}$ & \small $11.58\pm0.29$ & \small $11.73\pm0.30$ & \small $10.89\pm0.39$ & \small $22.10\pm2.79$ & \small $9.47\pm0.48$ & \small $11.54\pm0.54$ & \small $10.88\pm0.71$ \\
\midrule
\multirow{3}{*}{Meps 19}
& \small Cover & \small $0.90\pm0.01$ & \small $0.90\pm0.01$ & \small $0.91\pm0.00$ & \small $0.90\pm0.01$ & \small $0.92\pm0.01$ & \small $0.90\pm0.01$ & \small $0.89\pm0.12$ & \small $0.90\pm0.01$ & \cellcolor[HTML]{C0C0C0}\small $0.90\pm0.01$ \\
& \small Cond\_Cover & \small $0.89\pm0.02$ & \small $0.90\pm0.03$ & \small $0.90\pm0.04$ & \small $0.86\pm0.03$ & \small $0.88\pm0.02$ & \small $0.83\pm0.03$ & \small $0.81\pm0.15$ & \small $0.85\pm0.03$ & \cellcolor[HTML]{C0C0C0}\small $0.87\pm0.04$ \\
& \small Eff & \small $30.75\pm1.42$ & \small $17.41\pm1.23$ & \small $19.35\pm1.47$ & \small $29.91\pm2.11$ & \small $23.73\pm1.03$ & \small $65.66\pm23.88$ & \small $22.74\pm1.24$ & \small $16.95\pm0.66$ & \cellcolor[HTML]{C0C0C0}\small $\mathbf{14.94\pm2.96}$ \\
\midrule
\multirow{3}{*}{Meps 20}
& \small Cover & \small $0.90\pm0.01$ & \small $0.90\pm0.01$ & \small $0.91\pm0.01$ & \small $0.90\pm0.01$ & \small $0.92\pm0.01$ & \small $0.91\pm0.02$ & \small $0.88\pm0.02$ & \small $0.90\pm0.02$ & \cellcolor[HTML]{C0C0C0}\small $0.90\pm0.01$ \\
& \small Cond\_Cover & \small $0.90\pm0.02$ & \small $0.90\pm0.04$ & \small $0.89\pm0.00$ & \small $0.90\pm0.03$ & \small $0.85\pm0.03$ & \small $0.84\pm0.04$ & \small $0.80\pm0.04$ & \small $0.85\pm0.03$ & \cellcolor[HTML]{C0C0C0}\small $0.85\pm0.03$ \\
& \small Eff & \small $28.27\pm1.76$ & \small $17.84\pm1.41$ & \small $18.73\pm1.72$ & \small $28.71\pm3.03$ & \small $23.81\pm1.22$ & \small $71.10\pm26.01$ & \small $19.46\pm1.59$ & \small $16.06\pm1.87$ & \cellcolor[HTML]{C0C0C0}\small $\mathbf{14.02\pm2.86}$ \\
\midrule
\multirow{3}{*}{Meps 21}
& \small Cover & \small $0.90\pm0.01$ & \small $0.90\pm0.01$ & \small $0.90\pm0.02$ & \small $0.90\pm0.01$ & \small $0.91\pm0.01$ & \small $0.90\pm0.01$ & \small $0.88\pm0.06$ & \small $0.90\pm0.01$ & \cellcolor[HTML]{C0C0C0}\small $0.90\pm0.02$ \\
& \small Cond\_Cover & \small $0.90\pm0.02$ & \small $0.90\pm0.03$ & \small $0.89\pm0.05$ & \small $0.89\pm0.02$ & \small $0.84\pm0.02$ & \small $0.83\pm0.03$ & \small $0.80\pm0.08$ & \small $0.86\pm0.03$ & \cellcolor[HTML]{C0C0C0}\small $0.82\pm0.04$ \\
& \small Eff & \small $30.42\pm1.33$ & \small $17.89\pm1.35$ & \small $20.23\pm2.26$ & \small $29.42\pm2.13$ & \small $24.39\pm3.17$ & \small $53.65\pm18.49$ & \small $19.22\pm1.60$ & \small $16.93\pm1.32$ & \cellcolor[HTML]{C0C0C0}\small $\mathbf{14.16\pm1.97}$ \\
\midrule
\multirow{3}{*}{Bike}
& \small Cover
& \small $0.90\pm0.01$ & \small $0.90\pm0.01$ & \small $0.91\pm0.01$ & \small $0.90\pm0.01$ & \small $0.90\pm0.01$ & \small $0.90\pm0.02$ & \small $0.90\pm0.01$ & \small $0.90\pm0.02$ & \cellcolor[HTML]{C0C0C0}\small $0.90\pm0.02$ \\
& \small Cond\_Cover
& \small $0.80\pm0.02$ & \small $0.84\pm0.03$ & \small $0.81\pm0.03$ & \small $0.81\pm0.03$ & \small $0.80\pm0.05$ & \small $0.82\pm0.04$ & \small $0.83\pm0.05$ & \small $0.80\pm0.05$ & \cellcolor[HTML]{C0C0C0}\small $0.81\pm0.05$ \\
& \small Eff ($\times 10^{-3}$)
& \small $0.54\pm0.02$ & \small $0.49\pm0.02$ & \small $0.44\pm0.02$ & \small $0.43\pm0.02$ & \small $0.49\pm0.03$ & \small $1.07\pm0.13$ & \small $0.79\pm0.09$ & \small $0.46\pm0.03$ & \cellcolor[HTML]{C0C0C0}\small $\mathbf{0.36\pm0.00}$ \\
\midrule
\multirow{3}{*}{Temperature}
& \small Cover & \small $0.90\pm0.02$ & \small $0.90\pm0.01$ & \small $0.90\pm0.01$ & \small $0.90\pm0.02$ & \small $0.90\pm0.01$ & \small $0.90\pm0.02$ & \small $0.90\pm0.02$ & \small $0.90\pm0.01$ & \cellcolor[HTML]{C0C0C0}\small $0.90\pm0.01$ \\
& \small Cond\_Cover & \small $0.89\pm0.04$ & \small $0.89\pm0.05$ & \small $0.88\pm0.03$ & \small $0.88\pm0.05$ & \small $0.85\pm0.03$ & \small $0.80\pm0.04$ & \small $0.79\pm0.04$ & \small $0.85\pm0.03$ & \cellcolor[HTML]{C0C0C0}\small $0.82\pm0.03$ \\
& \small Eff & \small $3.45\pm0.10$ & \small $3.42\pm0.12$ & \small $3.43\pm0.10$ & \small $3.46\pm0.11$ & \small $2.59\pm0.07$ & \small $3.78\pm0.11$ & \small $2.55\pm0.09$ & \small $3.80\pm0.01$ & \cellcolor[HTML]{C0C0C0}\small $\mathbf{2.43\pm0.13}$ \\
\bottomrule[1pt]
\end{tabular}
\begin{tablenotes}
\item[1] \footnotesize Metrics include marginal coverage (Cover), worst-slab conditional coverage (Cond\_Cover), and prediction set efficiency (Eff). Values report mean $\pm$ standard error over $30$ runs. The lowest efficiency per dataset is bolded and highlighted along with corresponding coverage metrics.
\end{tablenotes}
\end{threeparttable}}
\label{tab:real_data_summary_full}
\end{table*}

\section{Sensitivity Analysis}
\label{appendix:sensitivity_analysis}
This section provides additional sensitivity analyses to assess the robustness, scalability, and geometric flexibility of \texttt{ORCA}. We examine four practical factors: model misspecification in the fitted conditional generator, the choice of the $m$-nearest-neighbor ranking parameter, the runtime--efficiency tradeoff of exact and approximate Stage~2 optimization, and the role of the output-space metric in high-dimensional settings. Overall, the results show that \texttt{ORCA} is stable under moderate parameter choices, remains effective under reasonable generative-model misspecification, and can be implemented efficiently through the proposed approximation schemes.

\paragraph{Sensitivity to generative model misspecification.}
We evaluate \texttt{ORCA} robustness to misspecification of the conditional model $\hat p(y\mid x)$. We use Quantile Random Forests (QRF), whose model complexity is controlled by the minimum samples per leaf (``\#leaf''), an interpretable bias--variance knob for $\hat p(y\mid x)$. Table~\ref{tab:sens-misspec} reports \texttt{ORCA} efficiency and its ratio relative to PCP and CHR (smaller is better) on the S-shape dataset. As expected, performance degrades at the extremes: \#leaf$=1$ overfits and yields noisy density estimates, whereas very large \#leaf oversmooths and washes out density structure. \texttt{ORCA} performs best at moderate \#leaf, where $\hat p(y\mid x)$ captures the main structure with controlled variance.
\begin{table}[!tbh]
\centering
\caption{\textbf{Sensitivity to conditional model quality (QRF).} Varying \#leaf controls the quality of $\hat p(y\mid x)$. We report marginal coverage (Cov.), conditional coverage (CondCov.), \texttt{ORCA} efficiency (Eff.), and efficiency ratios relative to PCP/CHR (smaller is better).}
\footnotesize
\setlength{\tabcolsep}{3.2pt}
\renewcommand{\arraystretch}{0.95}
\begin{tabular}{lccccc}
\toprule
\#leaf & Cov. & CondCov. & Eff\ (\texttt{ORCA}) & $\frac{\mathrm{Eff(\texttt{ORCA})}}{\mathrm{Eff(PCP)}}$ & $\frac{\mathrm{Eff(\texttt{ORCA})}}{\mathrm{Eff(CHR)}}$ \\
\midrule
1   & 0.90 & 0.86 & 3.99 & 1.00 & 1.14 \\
20  & 0.89 & 0.84 & 0.11 & 0.84 & 0.03 \\
50  & 0.90 & 0.85 & 0.21 & 0.87 & 0.06 \\
100 & 0.90 & 0.87 & 0.52 & 0.85 & 0.16 \\
200 & 0.90 & 0.85 & 0.77 & 1.03 & 0.22 \\
\bottomrule
\end{tabular}

\vspace{-7pt}
\label{tab:sens-misspec}
\end{table}

\vspace{-1pt}
\paragraph{Sensitivity to $m$-NN ranking parameter.}
\texttt{ORCA} ranks candidates using $m$-nearest-neighbor statistic over the conditional samples. Small $m$ yields high-variance rankings, while larger $m$ smooths local geometry and stabilizes the ordering. We vary $m$ and report the fractional efficiency gain over PCP $\frac{\widehat{\mathrm{Eff}}(\mathrm{PCP})-\widehat{\mathrm{Eff}}(\mathrm{\texttt{ORCA}})} {\widehat{\mathrm{Eff}}(\mathrm{PCP})},$
where positive values indicate smaller sets than PCP. Table~\ref{tab:sens-m} shows consistent gains for $m\ge 7$, whereas very small $m$ can lead to unstable ranks and weaker improvements. Overall, our method performs well for moderate choices of $m$.

\begin{table}[!tbh]
\centering
\caption{\textbf{Sensitivity to $m$-NN ranking.} \texttt{ORCA} efficiency improvement over PCP as a function of the $m$-NN parameter; moderate $m$ stabilizes ranking and yields consistent gains.}
\footnotesize
\setlength{\tabcolsep}{2.6pt}
\renewcommand{\arraystretch}{0.95}
\begin{tabular}{lrrrrrrr}
\toprule
& \multicolumn{7}{c}{Number of neighbors $m$} \\
\cmidrule(lr){2-8}
Method & 1 & 4 & 7 & 10 & 13 & 16 & 19 \\
\midrule
\texttt{ORCA}-MILP & -0.022 & -0.199 & \textbf{0.264} & \textbf{0.318} & \textbf{0.316} & \textbf{0.289} & \textbf{0.279} \\
\texttt{ORCA}-LP   & -0.727 & -1.131 & \textbf{0.159} & \textbf{0.229} & \textbf{0.269} & \textbf{0.293} & \textbf{0.275} \\
\bottomrule
\end{tabular}

\vspace{-5pt}
\label{tab:sens-m}
\end{table}

\paragraph{Runtime--efficiency tradeoff.}
We compare the runtime and efficiency of PCP, \texttt{ORCA}-MILP, \texttt{ORCA}-LP, and \texttt{ORCA}-Greedy on the S-shape dataset. Table~\ref{tab:runtime-eff} reports the mean runtime in seconds and prediction-set efficiency over 10 independent runs, where smaller efficiency values indicate tighter prediction sets. Overall, \texttt{ORCA}-Greedy provides the most favorable practical tradeoff: it is orders of magnitude faster than the exact MILP while achieving nearly identical, and in some cases slightly better, efficiency. For example, when $K=20$, \texttt{ORCA}-Greedy matches the efficiency of \texttt{ORCA}-MILP while reducing runtime from $23.15$ seconds to $0.06$ seconds. \texttt{ORCA}-LP also substantially reduces runtime relative to the MILP and maintains competitive efficiency, though it is consistently slower than the greedy variant. \texttt{ORCA}-Naive is substantially slower and is included only as a reference implementation. These results suggest that the greedy approximation is a scalable and effective default choice for practical deployments of \texttt{ORCA}.
\begin{table}[!tbh]
\centering
\caption{\textbf{Runtime and efficiency vs.\ $K$.}}
\footnotesize
\setlength{\tabcolsep}{1.5pt}
\renewcommand{\arraystretch}{0.8}
\begin{tabular}{lcccc}
\toprule
 & \multicolumn{4}{c}{time (s) / Eff.} \\
\cmidrule(lr){2-5}
Method & $K{=}10$ & $15$ & $20$ & $25$ \\
\midrule
PCP         & 0.00/0.75 & 0.00/0.48 & 0.00/0.37 & 0.00/0.34 \\
\texttt{ORCA}-MILP   & 11.95/0.67 & 16.97/0.35 & 23.15/0.32 & 40.96/0.31 \\
\texttt{ORCA}-LP     & 1.50/0.67 & 2.88/0.37 & 5.22/0.34 & 8.94/0.36 \\
\texttt{ORCA}-Greedy & 0.02/0.66 & 0.03/0.35 & 0.06/0.32 & 0.09/0.32 \\
\texttt{ORCA}-Naive  & 174.12/0.69 & 442.16/0.37 & 548.92/0.35 & 665.89/0.33 \\
\bottomrule
\end{tabular}

\label{tab:runtime-eff}
\vspace{-8pt}
\end{table}

\paragraph{Geometry and metric choice.}
Methodologically, \texttt{ORCA} is not restricted to Euclidean distance. For any valid distance $d(\cdot,\cdot)$ on the output space, the same rank-adaptive construction applies directly, and the finite-sample conformal coverage guarantee remains unchanged. More generally, \texttt{ORCA} can be viewed as a coverage-allocation framework over ranked generated samples, where the geometry of the prediction set is determined by the chosen output-space metric.
\begin{table}[!tbh]
\centering
\caption{\textbf{Metric ablation on MNIST-VAE.} We compare PCP and \texttt{ORCA} under different output-space metrics. Coverage is reported together with the off-class rate (OCR), where smaller OCR indicates fewer semantically incorrect images included in the prediction set.}
\label{tab:metric_ablation}
\footnotesize
\setlength{\tabcolsep}{5pt}
\renewcommand{\arraystretch}{0.9}
\begin{tabular}{lcccc}
\toprule
Metric
& PCP Cover
& PCP OCR
& \texttt{ORCA} Cover
& \texttt{ORCA} OCR \\
\midrule
Pixel-space $\ell_2$
& 0.899 & 0.374 & 0.900 & \textbf{0.338} \\
Latent-space $\ell_2$
& 0.909 & 0.178 & 0.908 & \textbf{0.123} \\
Latent cosine distance
& 0.906 & 0.090 & 0.897 & \textbf{0.054} \\
Learned latent metric
& 0.901 & 0.062 & 0.905 & \textbf{0.019} \\
\bottomrule
\end{tabular}
\vspace{-6pt}
\end{table}

To examine this flexibility, we extend the MNIST-VAE experiment using four metrics: pixel-space $\ell_2$ distance, latent-space $\ell_2$ distance with latent dimension $16$, latent-space cosine distance, and a learned latent metric. Since raw prediction-set size is not directly comparable across different metrics, we instead report the \emph{off-class rate} (OCR). For a target input digit class $c$, OCR measures the probability that images from classes other than $c$ are included in the uncertainty set. A smaller OCR therefore indicates that the prediction set contains fewer semantically incorrect images. As shown in Table~\ref{tab:metric_ablation}, both PCP and \texttt{ORCA} achieve coverage close to the nominal level, while \texttt{ORCA} consistently attains lower OCR across all metrics. The improvement is especially pronounced for learned latent metrics, suggesting that incorporating task-relevant geometry can substantially improve the semantic quality of conformal prediction sets in high-dimensional output spaces.

\end{document}